%% file: main.tex
\documentclass[sort,compress,11pt]{elsarticle}

\usepackage{tikz}
\usetikzlibrary{arrows.meta,positioning,fit,calc,shapes.misc}
\usepackage{amssymb}
\usepackage{amsthm}
\usepackage{amsmath}
\usepackage{mathtools}
\usepackage{mathrsfs}
\usepackage[ruled, lined, linesnumbered, longend]{algorithm2e}
\SetKw{init}{Initialization:}
\let\oldnl\nl% Store \nl in \oldnl
\newcommand{\nonl}{\renewcommand{\nl}{\let\nl\oldnl}}% Remove line number for one line
\usepackage{array}
\usepackage{multirow}
\usepackage{listings}
\usepackage{tabu}
\usepackage{enumerate}
\usepackage{lineno}
\usepackage{fullpage}
\usepackage{xcolor}
\usepackage[colorinlistoftodos]{todonotes}
\usepackage{bbm}
\usepackage[colorlinks=true]{hyperref}
\usepackage{url}
\usepackage{textcomp}
\usepackage{gensymb}
\usepackage{soul}
\usepackage{graphicx}
\usepackage{subfig}
\usepackage{array}
\usepackage{makecell}
\usepackage[export]{adjustbox}
\usepackage{float}
\usepackage{caption}
\usetikzlibrary{shapes}
\biboptions{numbers,comma,round,square}
\usepackage{arydshln}
\usepackage[para,online,flushleft]{threeparttable}
\usepackage{orcidlink}

\definecolor{myred}{RGB}{214,39,40}
\definecolor{mygray}{RGB}{176,176,176}
\definecolor{myorange}{RGB}{255,127,14}
\definecolor{mygreen}{RGB}{44,160,44}
\definecolor{mylightgray}{RGB}{204,204,204}
\definecolor{mypurple}{RGB}{148,103,189}
\definecolor{mybrown}{RGB}{140,86,75}
\definecolor{steelblue}{RGB}{31,119,180}
\definecolor{intraray}{RGB}{127,193,219}
\definecolor{hemitrichs}{RGB}{26,74,93}

\theoremstyle{definition}

\theoremstyle{remark}

\graphicspath{ {./figs/} }

\journal{Elsevier}
\DontPrintSemicolon
\definecolor{orcidlogocol}{HTML}{A6CE39}
\begin{document}
\makeatletter
\def\ps@pprintTitle{%
  \let\@oddhead\@empty
  \let\@evenhead\@empty
  \let\@oddfoot\@empty
  \let\@evenfoot\@oddfoot
}
\makeatother
\begin{frontmatter}

\title{Offline Reinforcement Learning for Fluid Controls: Data-based Multi-observational Policy Extraction}

\author[ndAME]{Deepak Akhare}%\corref{contrib}}
\author[llnl]{Luning Sun}%\corref{contrib}}
\author[ndAME]{Xin-Yang Liu}%\corref{contrib}}
\author[ndAME]{Xiantao Fan}%\corref{contrib}}
\author[llnl]{Timo Bremer}
\author[llnl]{Ben Zhu}%\corref{contrib}}
\author[ndAME, cornell]{Jian-Xun Wang\corref{corxh}}%\corref{contrib}}
% \author[ndAME]{Xin-Yang Liu\orcidlink{0000-0003-1423-605X}}%\corref{contrib}}

\address[ndAME]{Department of Aerospace and Mechanical Engineering, University of Notre Dame, Notre Dame, IN}
\address[llnl]{Lawrence Livermore National Laboratory, University of Notre Dame, Livermore, CA}
\address[cornell]{Sibley School of Mechanical and Aerospace Engineering, Cornell University, Ithaca, NY, USA}

% \cortext[contrib]{Authors contributed equally}
% \cortext[corxh]{Corresponding author. Tel: +1 574-631-5302}
% \ead{xliu28@nd.edu}

\begin{abstract}
Active flow control is a fundamental application in engineering. Recent advances in deep reinforcement learning have made progress in this field. However, the classical online RL approaches require extensive real-time interactions with the high fidelity environment, while each sensor configuration change necessitates whole policy retraining. All these factors result in prohibitive computational costs for real-world applications. In this work, we propose a novel offline RL framework that addresses both challenges through data-driven policy extraction. We develop a sensor position-conditioned architecture that enables a single policy network to adapt seamlessly to multiple sensor arrangements. The position-conditioned approach incorporated spatial relationship modeling through Point Attention layers to ensure the generalizability to varying sensor placements. We demonstrate the framework on two representative problems, mitigating chaoticity in the Kuramoto-Sivashinsky equation and flow control over airfoils governed by the Navier-Stokes equation. The result demonstrates that the policy extraction from the dataset provides unprecedented flexibility for sensor placement optimization. This approach represents a significant step towards adaptive, intelligent flow control systems.
\end{abstract}

\begin{keyword}
Active Flow Control \sep Offline Reinforcement Learning \sep Sensor Placement \sep Policy Extraction \sep Ensemble Methods
\end{keyword}
\end{frontmatter}

\section{Introduction}
\label{sec:Intro}

% Offline RL and its rise
% Offline Rl in fluids 
% policy extraction
% Multi-observational Policy

Flow control is a ubiquitous and critical challenge across a wide range of engineering and scientific applications, including aerodynamics~\cite{glezer2002synthetic,liu2024deep,patel2024enhancing,vinuesa2022flow,zhao2025optimal}, wind energy~\cite{aubrun2017review, simley2024wake,xie2023data,fernandez2022actor,dong2021intelligent,dallas2024control}, drag reduction~\cite{garcia2011review, garcia2025deep,sonoda2023reinforcement,xia2024active,guastoni2023deep,suarez2024active}, noise mitigation in propulsion systems~\cite{maceda2023stabilization}, enhanced mixing in chemical processes~\cite{ottino1990mixing,vignon2023effective}, fluid–structure interaction (FSI)~\cite{paidoussis2010fluid,chen2023deep,yao2024deep}, and soft robotics~\cite{rus2015design,verma2018efficient}. The economic impact is substantial, with drag reduction alone potentially saving billions in fuel costs annually across transportation sectors.

 In 1982, Liepmann et al achieved the first successful demonstration on the cancellation of laminar-instability waves~\cite{liepmann1982control}.
Bushnell et al's comprehensive review paper~\cite{bushnell1989turbulence} shows various methods for turbulence control and highlights their sensitivity to various inputs that influence flow structures.  Joshi et al.~\cite{joshi1997control} and Bewley et al.~\cite{bewley1998optimal} were pioneers in applying modern control theory to fluid dynamics, developing LQG and robust $\mathcal{H}_\infty$ control frameworks. However, the underlying assumption of linearity limited the applicability of these methods to highly nonlinear transitional or turbulent flows. These classical approaches faced fundamental computational barriers: matrix operations requiring expensive O(n³) inversions, optimization times measured in hours or days that precluded real-time applications, and combinatorial sensor placement problems that became intractable beyond 20-30 sensor locations.

Recently, researchers have also conducted many relevant studies on fluid control and optimal sensor placement using a deep reinforcement learning based method (DRL). It has emerged as a powerful tool for solving complex decision-making problems by learning optimal policies through interactions with an environment. Paris et al.~\cite{paris2021robust} proposed a variant of the Proximal Policy Optimization (PPO) algorithm for optimal sensor placement. They leverage the sparsity-promoting loss to learn from a pretrained policy given the sparse sensor constraints. Watanabe et al ~\cite{watanabe2025effect} investigated the effect of pressure sensor placement for the experimental deep stall control by reinforcement learning. 
% Their result manifested that installing sensors around the leading edge is the most effective layout. 
However, conventional RL approaches typically require extensive online interactions with the environment during training, which can be impractical, expensive, or even unsafe in real-world applications such as robotics, healthcare, and industrial processes. The computational cost scales poorly: online DRL typically requires $10^4-10^6$ environment interactions, translating to weeks or months of real-time fluid experiments, making practical deployment prohibitively expensive. These limitations significantly restrict the widespread applicability and scalability of traditional RL methodologies. One promising direction to reduce the cost of online DRL is to learn a fast surrogate model. A common practice is to interact with low-fidelity (LF) simulation environment to simplify the problem and get rid of the expensive high-fidelity (HF) environment~\cite{gazzola2016learning,novati2017synchronisation,gustavsson2017finding,colabrese2017flow}. Another promising way is to construct a neural network-based surrogate for DRL training. Liu et al~\cite{liu2021physics} developed a physics-informed surrogate model to incorporate PDE constraints into model-based RL training. Recent work also developed more advanced surrogate models leveraging differentiable solvers with learnable terms~\cite{shankar2023differentiable,liu2024multi,fan2025diff} and applied it to the control task~\cite{sun2025multi}. Bae et al~\cite{bae2022scientific} also proposed to leverage multi-agent RL to reduce the computational cost by several orders.  

% There is also emerging research on leveraging generative models for sensor placement. Vinuesa et al~\cite{vishwasrao2025diff} proposed the Diff-SPORT framework, where a diffusion model is trained as a prior and a shapley-based sensor attribution is used to identify compact, spatially contigious subregions for sensor placement in urban flows. However, there are several revenues remain to improve the generalizability and lower the computational cost for the diffusion-based framework.  

Alternatively, offline reinforcement learning (offline RL) offers a promising solution by enabling the learning of effective control policies directly from pre-collected datasets, eliminating the need for online environmental interactions~\cite{levine2020offline, prudencio2023survey, kumar2020conservative}. By leveraging historical data collected from expert demonstrations, prior training runs, or heuristic-based policies, offline RL facilitates efficient and safer policy development, reducing the costs and risks associated with live environment interaction. Modern offline RL algorithms have addressed key challenges such as distributional shift through conservative estimation methods~\cite{kumar2019stabilizing,wu2019behavior}, behavioral cloning~\cite{fujimoto2019off}, and uncertainty quantification~\cite{an2021uncertainty}. This makes offline RL especially valuable in sensitive, resource-constrained, or high-stakes environments. Offline RL has demonstrated success in robotics manipulation tasks, achieving over 90\% success rates from static datasets~\cite{singh2020end,mandlekar2021learning}, and in healthcare applications where online experimentation is ethically constrained~\cite{gottesman2019guidelines}, and autonomous driving where safety-critical constraints prohibit extensive online exploration~\cite{kiran2021deep}. However, this direction is underexplored in scientific machine leraning field. The most related work is the Char et al~\cite{char2023offline}, who applied offline model-based RL to tokamak control using historical data from DIII-D fusion device.

In the fluid control task, there is still one particular challenge regarding reliance on partial observations obtained through fixed sensor configurations. In practical deployments, sensor locations may need to change due to experimental constraints, hardware limitations, or new objectives. However, most policies are trained assuming a specific observation setup, making them highly sensitive to any modification in sensor placement. Even slight changes in the sensor configuration typically necessitate retraining the policy, which may involve further data collection or online interactions. This creates a critical bottleneck: each sensor configuration change requires a separate $O(n^2)$ policy optimization, making adaptive sensing computationally intractable for real-time applications.

In this work, we explore the potential of offline RL—specifically using the SACN~\cite{kumar2019stabilizing} algorithm—to extract multiple distinct policies from a single dataset. We demonstrate that once data is collected, whether from human demonstrations or learned behaviour, it can be reused to train policies for different observation settings, such as varying sensor placements. This approach eliminates the need for interactions with the environment and enables flexible adaptation of control strategies across a range of configurations using the same dataset.

However, training a separate policy for every slight variation in sensor location remains inefficient and time-consuming. The computational complexity grows linearly with the number of sensor configurations, creating an $O(k\times n^2)$ scaling problem where k represents the number of different sensing configurations. To address this, we propose a sensor position-conditioned policy network along with a training strategy that enables a single network to generalize across multiple sensor configurations. Our approach leverages Point Attention architecture to capture spatial relationships between sensors, achieving permutation invariance and enabling generalization across arbitrary sensor layouts. This reduces the computational burden from $O(k\times n^2)$ to $O(n^2)$ while maintaining comparable performance across all configurations. This significantly improves the practicality of offline RL by allowing seamless adaptation to sensor changes during deployment, without requiring additional retraining or data collection.
The paper is organized as the following: Sec.~\ref{subsec: problem} to ~\ref{subsec: offline_SAC_N} formulate the problem setting and review the relevant RL background. And Sec.~\ref{subsec:SApi} elaborates our contribution on Sensor-ware Policy Network (PC$\pi$-net). Sec.~\ref{subsec:test_env} presents the fluid dynamics cases in current work. We analyze the performance of our proposed framework in Sec.~\ref{sec:Result}. Sec.~\ref{sec:Discussion} discusses the sensor optimization and Sec.~\ref{sec:Conclusion} concludes the paper.

\input{./method}

\input{./result}
% \newpage
\input{./discussion}

\input{./conclusion}

\section*{Acknowledgements}
This work was performed under the auspices of the U.S. Department of Energy by the Lawrence Livermore National Laboratory under Contract No. DE-AC52-07NA27344.  We acknowledge support from LLNL Laboratory548
Directed Research and Development (LDRD) Program Grant No. 22-SI-004. The work is reviewed and
released under LLNL-JRNL-2012764.

\bibliographystyle{elsarticle-num}
\bibliography{ref}
\newpage
\appendix

\section{Various models}

% \subsection{SACN}
\begin{figure}[!ht]
    \centering
    \includegraphics[width=0.8\linewidth]{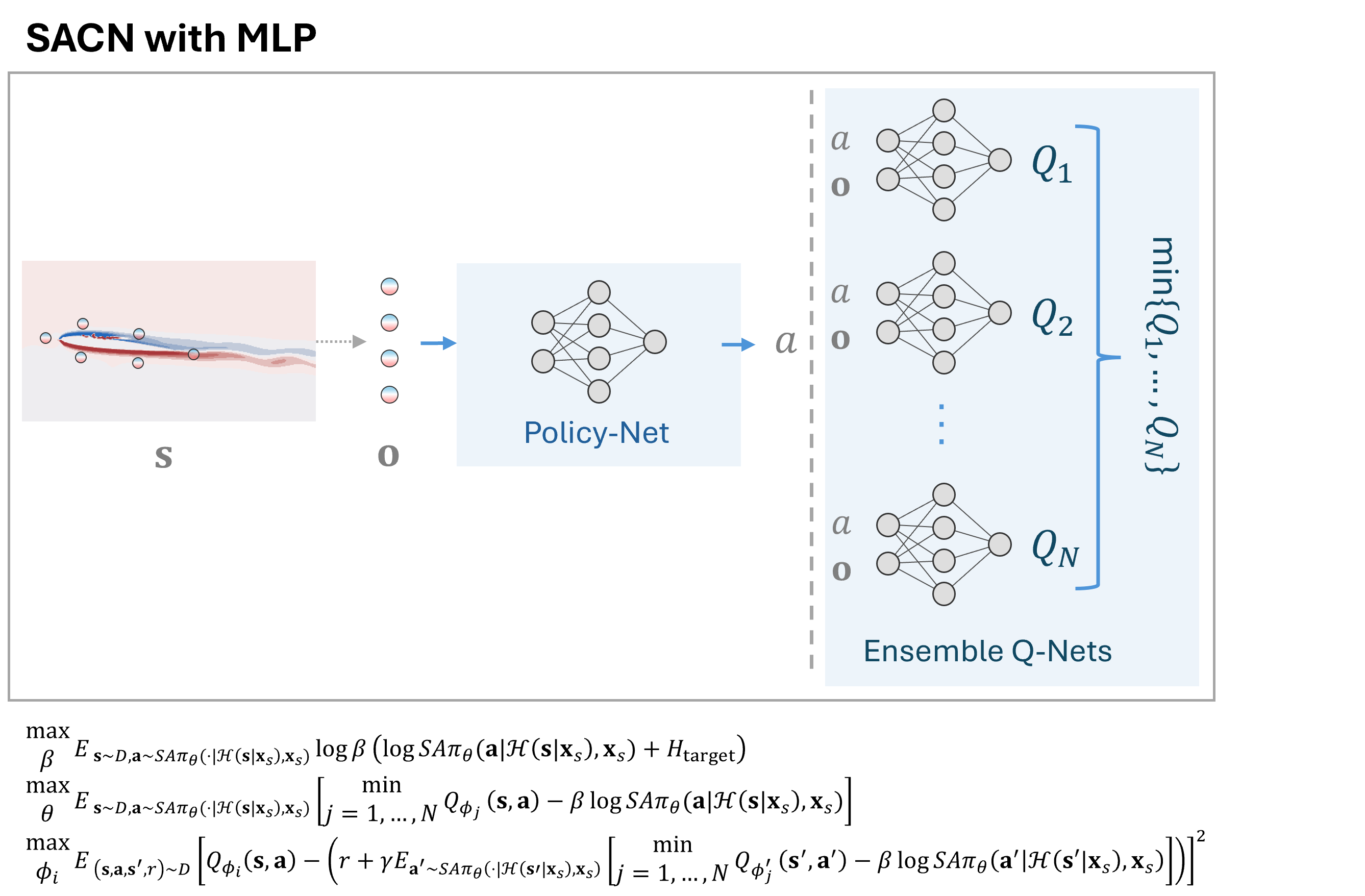}
    \caption{SACN algorithm with MLP as actor}
    \label{fig:sacn-mlp}
\end{figure}

% \subsection{SACN with PC$\pi$-net}
\begin{figure}[!ht]
    \centering
    \includegraphics[width=0.8\linewidth]{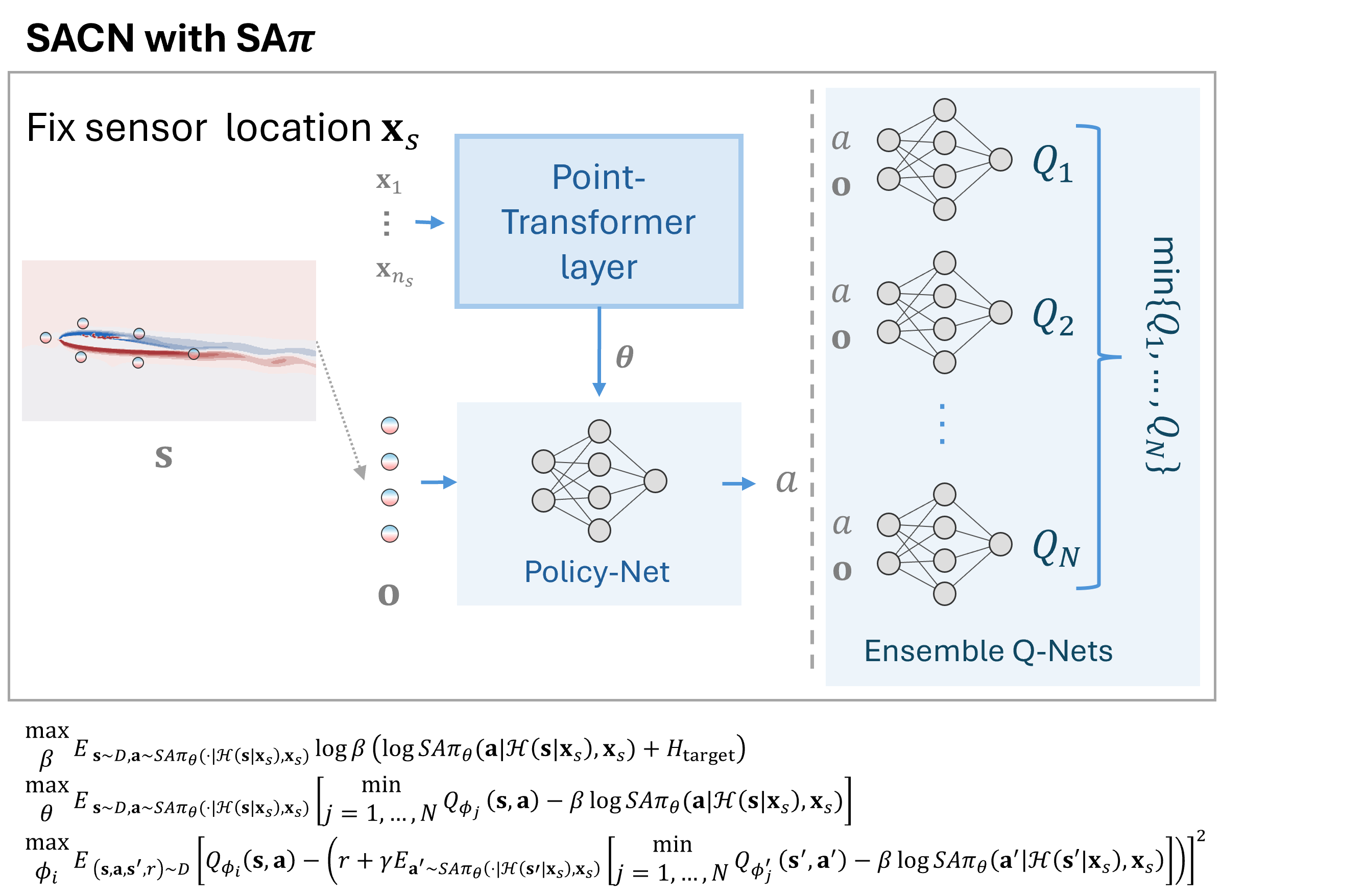}
    \caption{SACN algorithm with PC$\pi$-net as actor}
    \label{fig:sacn-sapi}
\end{figure}

% \subsection{PC-SACN with PC$\pi$-net}
\begin{figure}[!ht]
    \centering
    \includegraphics[width=0.8\linewidth]{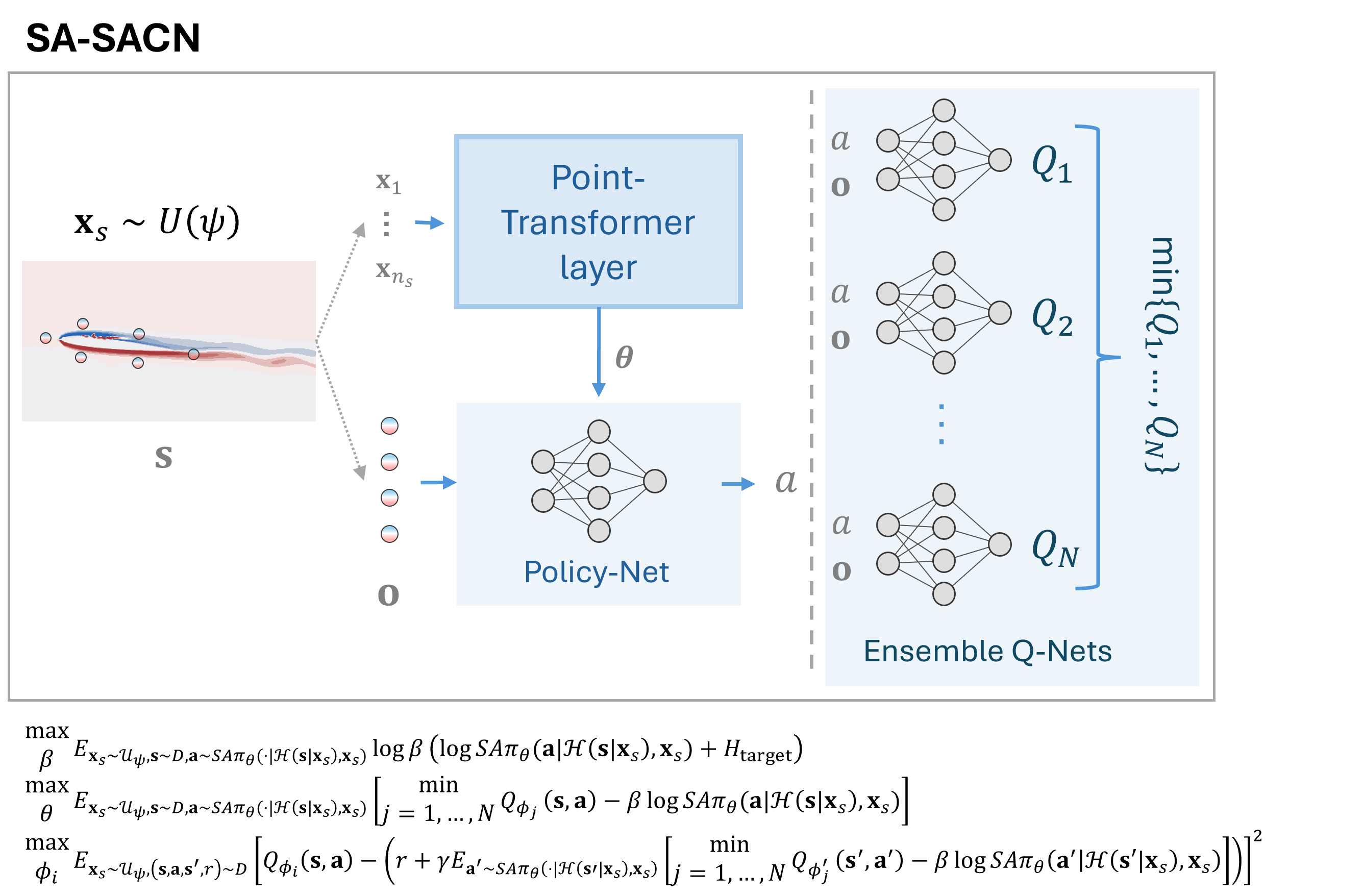}
    \caption{Sensor aware SACN (PC-SACN) algorithm with PC$\pi$-net as actor}
    \label{fig:PC-SACN}
\end{figure}

% \subsection{PC-SACN-opt with PC$\pi$-net}
\begin{figure}[!ht]
    \centering
    \includegraphics[width=0.8\linewidth]{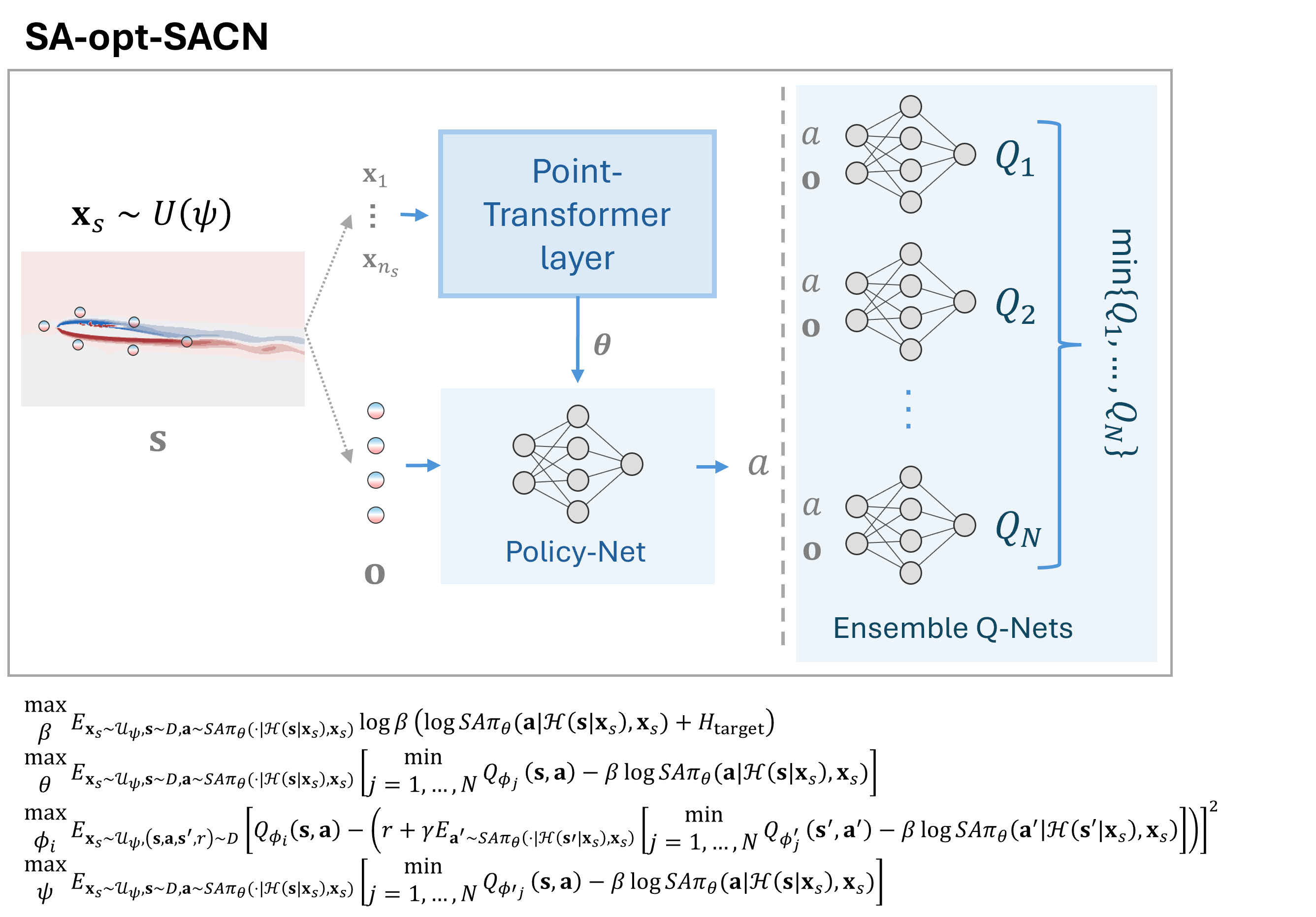}
    \caption{Sensor aware with optimization SACN (PC-SACN-opt) algorithm with PC$\pi$-net as actor}
    \label{fig:PC-SACN-opt}
\end{figure}

\section{Test on various policy data sets}
Initially, we tested the SACN algorithm on KS and flow over airfoil environments, with the goal of extracting an equal or better policy than the one present in the dataset. 

\begin{figure}[!htp]
    \centering
    \includegraphics[width=0.9\linewidth]{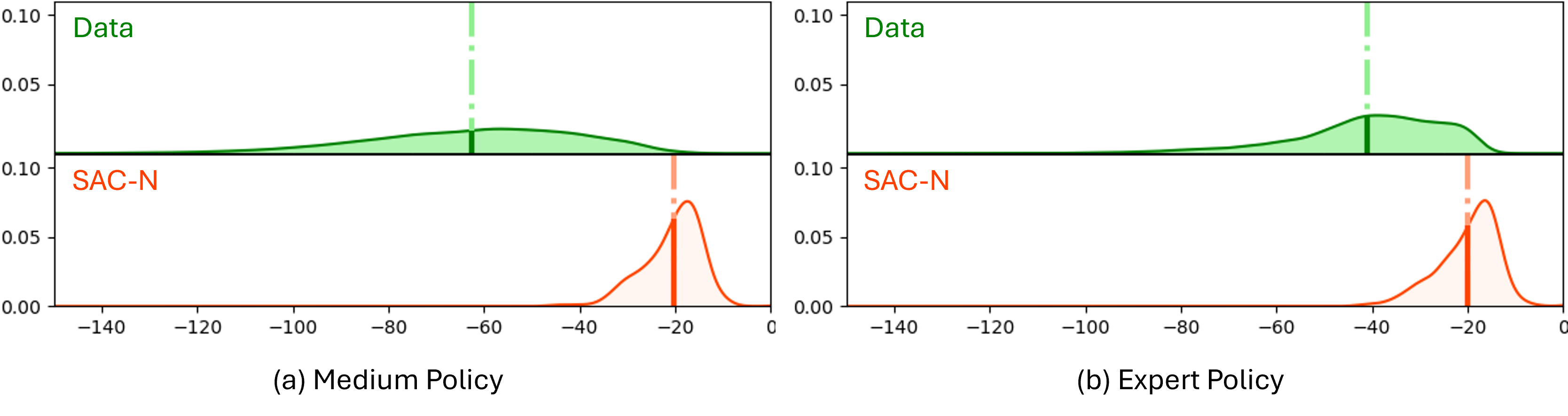}
    \caption{Comparison of return distributions between the learned policies using SACN and the various behavior policies used to generate the dataset in the KS environment. Policies are trained using SACN with MLP as policy Net.}
    \label{fig:ks_sacn_vdata}
\end{figure}

For the KS environment, we constructed two datasets—one from a medium policy and one from an expert policy—each comprising 10,000 trajectories of 400 control steps. Figure \ref{fig:ks_sacn_vdata} presents histograms of returns achieved by deploying the SACN–trained policy across 200 test rollouts. These results clearly demonstrate that the policy extracted by SACN significantly outperforms the original medium and expert policies used to generate the data.

\begin{figure}[!htp]
    \centering
    \includegraphics[width=0.9\linewidth]{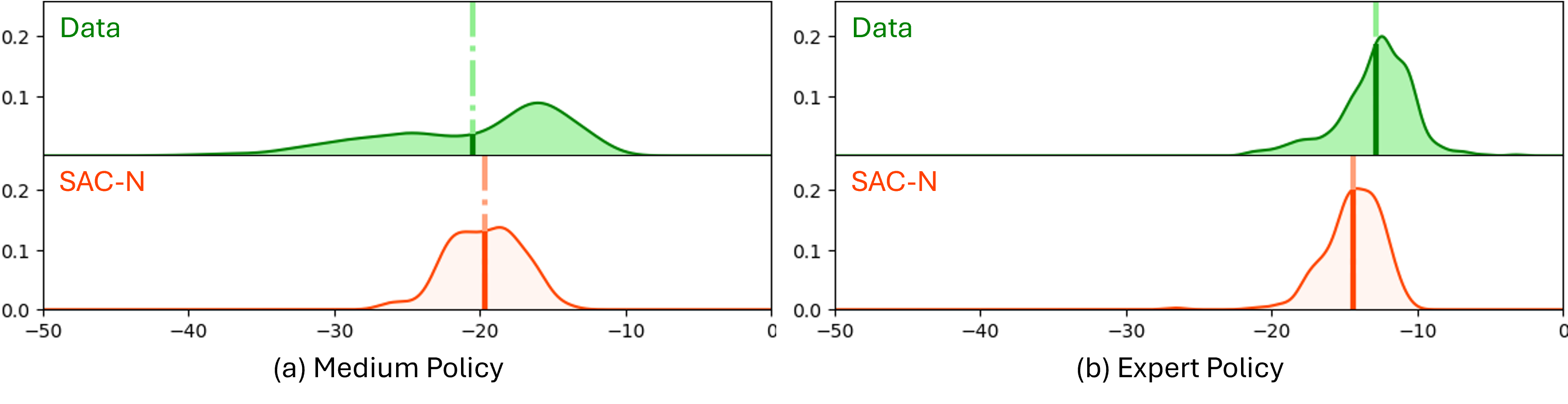}
    \caption{Comparison of return distributions between the learned policies using SACN and the various behavior policies used to generate the dataset in the flow over airfoil environment. Policies are trained using SACN with MLP as policy Net.}
    \label{fig:af_sacn_po_vdata}
\end{figure}

For the airfoil‐flow environment, we similarly built two datasets—one generated by a medium policy and one by an expert policy—each containing 2,000 trajectories of 400 control steps. Figure \ref{fig:af_sacn_po_vdata} shows histograms of returns from deploying the SACN–trained policy over 200 test rollouts. The results indicate that the SACN–extracted policy achieves performance on par with the original medium and expert policies used for data generation. This clearly demonstrates that the SACN can extract a policy from data that performs as well as or better than the behaviours inherently present in the dataset.

\newpage

\section{Details on KS environment}
\label{sec:KS_env}

The KS equation is a one-dimensional nonlinear partial differential equation given by:
\begin{equation}
    \frac{\partial u}{\partial t} + \frac{\partial^2 u}{\partial x^2} + \frac{\partial^4 u}{\partial x^4} + \frac{1}{2} u\frac{\partial u}{\partial x} = 0, \ \    x \in [0,l], \ t\in[0,\infty).
\end{equation}
This fourth-order equation is known for its ability to generate rich spatiotemporal chaotic behavior, making it a canonical model for studying turbulence in low-dimensional settings. To solve the KS equation numerically, we use a Fourier spectral method for spatial discretization, which exploits the periodic boundary conditions to compute derivatives efficiently in spectral space via Fast Fourier Transforms (FFTs). Time integration is performed using a semi-implicit third-order Runge–Kutta (RK3) scheme. 

In our experiments, we use a spatial grid of 64 points over a domain of length 4. The time step for integration is set to $\Delta t = 0.01$ sec, and the control action is applied at every 25 integration steps, corresponding to a control interval of 0.25 seconds.

\section{Details on flow over airfoil environment}
\label{sec:airfoil_env}
The flow over the airfoil is simulated using our in-house GPU-accelerated solver~\cite{fan2025diff}. The two-dimensional incompressible fluid motion is governed by the Navier–Stokes equations:

\begin{equation}
\label{eq:ns}
\begin{aligned}
\nabla\cdot\mathbf{u} = 0, \hspace{7em}&\mathbf{x}, t \in \Omega_f \times [0, T]\\
\frac{\partial{\mathbf{u}}}{\partial t} = -(\mathbf{u}\cdot\nabla){\mathbf{u}}+\nu\nabla^2\mathbf{u}-\frac{1}{\rho}\nabla{p}+\mathbf{f}_s, \hspace{3em} &\mathbf{x}, t \in \Omega_f \times [0, T]
\end{aligned} 
\end{equation}
where $t$ and $\mathbf{x}$ denote the temporal and spatial coordinates in the Eulerian frame. The fluid velocity $\mathbf{u}(\mathbf{x}, t)$ and pressure $p(\mathbf{x}, t)$ are defined over the fluid domain $\Omega_f \subset \mathbb{R}^2$. The parameters $\rho$ and $\nu$ represent the fluid density and kinematic viscosity, respectively. The external body force $\mathbf{f}_s$ accounts for interactions with immersed boundaries. The system is closed by appropriate initial and boundary conditions.

The fluid equations are discretized using a finite volume method (FVM) on a staggered Cartesian grid, where pressure is stored at cell centers and velocity components are located on corresponding cell faces. This arrangement enhances numerical stability and facilitates the enforcement of the incompressibility constraint. Convective terms are discretized using a first-order upwind scheme, while diffusive terms employ second-order central differences. Time integration is performed using an explicit first-order forward Euler method.

To enforce incompressibility, the Chorin projection method is employed. The immersed boundary method (IBM) is implemented via a direct-forcing approach, wherein a penalization force is applied to fluid cells intersecting the immersed structure. This force drives the local fluid velocity toward the prescribed solid velocity at the interface.

The solver has been validated against a range of two-dimensional fluid–structure interaction (FSI) benchmarks. Additional numerical details and validation results are provided in~\cite{fan2025diff}.

\begin{figure}[!hp]
    \centering
    \includegraphics[width=0.5\linewidth]{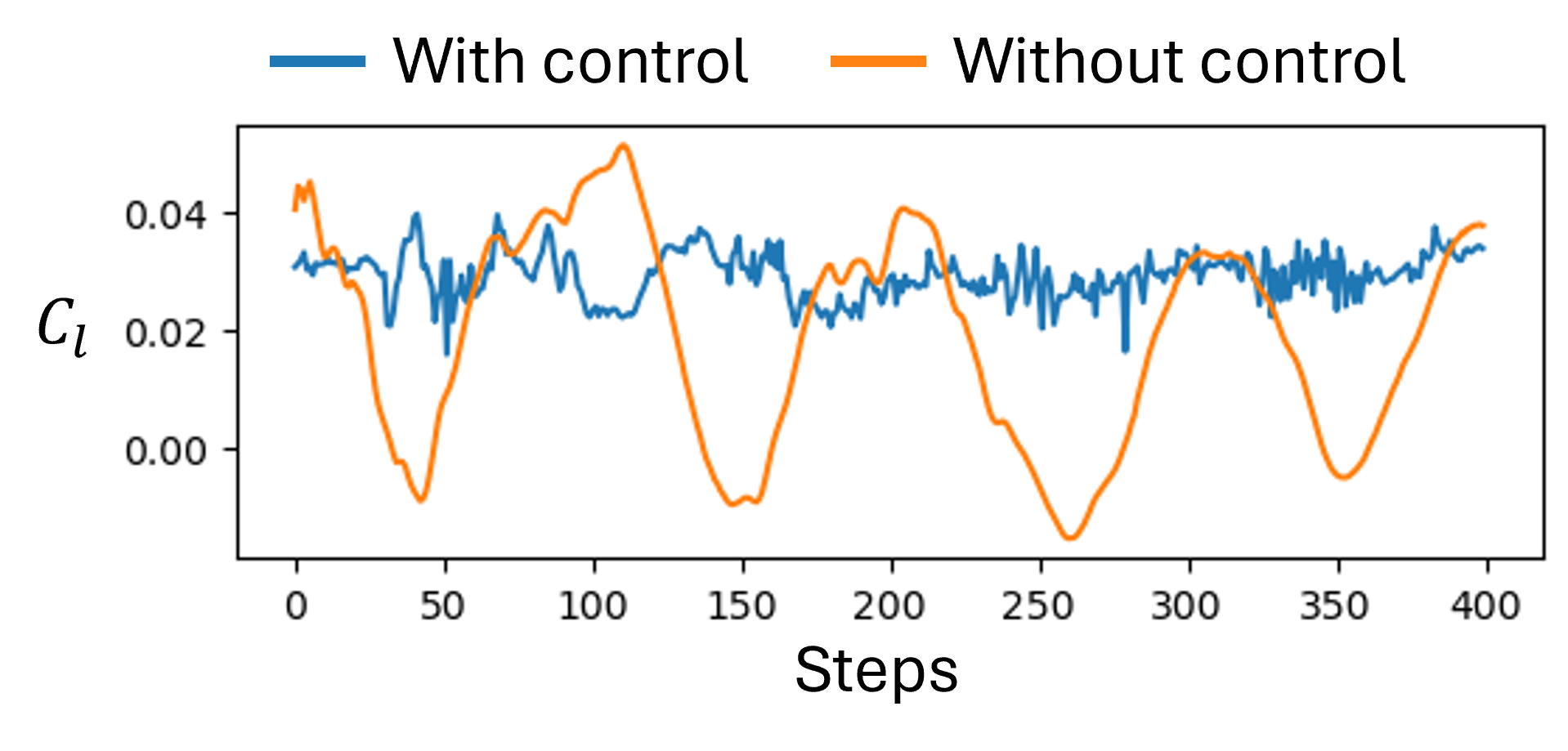}
    \caption{Comparison of lift coefficient $C_l$ with and without control applied. Jet control effectively reduces fluctuations in $C_l$.}
    \label{fig:Cl_airfoil}
\end{figure}

For the current study, we use a spatial grid of  $400 \times 400$ over a domain size of $20\times 20$.  A NACA0012 airfoil is placed in the domain with an angle of attack of 8 degrees. The time step for integration is set to $\Delta t = 0.01$ seconds, and control actions are applied every 10 integration steps, corresponding to a control interval of 0.1 seconds. The effect of the jet control on the $C_l$ is shown in Fig.~\ref{fig:Cl_airfoil}. Jet control effectively reduces fluctuations in $C_l$.

\section{Sensor Position-Conditioned SACN Algorithm}
\label{sec:PC-SACN-algo}

\begin{algorithm}[H]
\caption{Sensor Position-Conditioned SACN (PC-SACN)}
\label{alg:PC-SACN}
\KwIn{$D$: offline dataset; $\psi$: mean sensor layout; $\Delta$: perturbation; $N$: ensemble size; $\gamma$: discount; $\tau$: target rate; $\beta$: temperature; $\eta_Q,\eta_\pi,\eta_\psi$: learning rates; $B$: batch size}
\KwOut{Actor $\mathrm{PC}\pi_\theta$, critics $\{Q_{\phi_j}\}$}

Initialize $\theta, \{\phi_j\}_{j=1}^N, \{\phi'_j \!\leftarrow\! \phi_j\}$ and $\psi$\; % initialize networks and sensor layout

\For{each training step}{
    $\{(\mathbf{s},\mathbf{a},r,\mathbf{s}')\}_{b=1}^B \!\sim\! \mathcal{D}$ \tcp*{sample a batch of data from the dataset}
    $\mathbf{x}_s \gets \mathcal{U}_\psi$ \tcp*{sample sensor positions around $\psi$}
    
    \If{optimize sensors}{
        $\psi \leftarrow$ \text{OptimizeSensors}($\mathbf{s}, \mathbf{x}_s, \psi, \eta_\psi, \beta, B$) \tcp*{update mean sensor positions}
    }
    
    $\beta \leftarrow$ \text{UpdateBeta}($\mathbf{s}, \mathbf{x}_s, \eta_\beta, \beta, B$) \tcp*{adjust entropy temperature}
    
    $\theta \leftarrow$ \text{UpdateActor}($\mathbf{s}, \mathbf{x}_s, \theta, \eta_\theta, \beta, B$) \tcp*{update actor parameters}
    
    $\{\phi_j\}_{j=1}^{N} \leftarrow$ \text{UpdateQ}($\mathbf{s}, \mathbf{a}, r, \mathbf{s}' , \mathbf{x}_s, \{\phi_j\}_{j=1}^{N}, \eta_\beta, \beta, B$) \tcp*{update critic ensemble}
    
    $\phi'_j \leftarrow \tau \phi_j + (1-\tau)\phi'_j,\; \forall j$ \tcp*{soft update of target critics}
}
\end{algorithm}

\begin{algorithm}[H]
\caption{Sensor Position Optimization Function}\label{alg:secopt}
% \KwIn{$\mathbf{s}$: state, $\psi$: mean sensor positions, $\eta_\psi$: learning rates, $B$: batch size}
% \KwOut{Updated sensor positions $\psi$}
\SetKwFunction{FOptimize}{OptimizeSensors}
\SetKwProg{Fn}{Function}{:}{}

\Fn{\FOptimize{$\mathbf{s}, \mathbf{x}_s, \psi, \eta_\psi, \beta, B$}}{
    % $\mathbf{x}_s \gets \mathcal{U}_\psi$ \tcp*{sample sensor positions}\
    $\mathbf{o} \gets \mathcal{H}(\mathbf{s}\mid \mathbf{x}_s)$ \tcp*{compute partial observation}\
    $(\mathbf{a}, \log \pi_{\mathbf{a}}) \gets \mathrm{PC}\pi_\theta(\cdot \mid \mathbf{o}, \mathbf{x}_s)$ \tcp*{sample action, log-prob}\
    $\psi \gets \psi - \eta_\psi \nabla_{\psi} \frac{1}{B}\!\left[\beta \log \pi_{\mathbf{a}} - \min_j Q_{\phi_j}(\mathbf{s}, \mathbf{a})\right]$ \tcp*{update mean sensors positions}\
    \KwRet{$\psi$}\;
}
\end{algorithm}

\begin{algorithm}[H]
\caption{Beta Update Function}\label{alg:alphaopt}
\SetKwFunction{FOptimize}{UpdateBeta}
\SetKwProg{Fn}{Function}{:}{}

\Fn{\FOptimize{$\mathbf{s}, \mathbf{x}_s, \eta_\beta, \beta, B$}}{
    $\mathbf{o} \gets \mathcal{H}(\mathbf{s}\mid \mathbf{x}_s)$ \tcp*{compute partial observation}\
    $(\mathbf{a}, \log \pi_{\mathbf{a}}) \gets \mathrm{PC}\pi_\theta(\cdot \mid \mathbf{o}, \mathbf{x}_s)$ \tcp*{sample action, log-prob}\
    $\beta \gets \beta - \eta_\beta \nabla_{\beta} \frac{1}{B}\!\left[-\log\beta (\log \pi_{\mathbf{a}} + H_{\text{target}})\right]$ \tcp*{adjust entropy temperature}\
    \KwRet{$\beta$}\;
}
\end{algorithm}

\begin{algorithm}[H]
\caption{Actor Update Function}\label{alg:alphaopt}
\SetKwFunction{FOptimize}{UpdateActor}
\SetKwProg{Fn}{Function}{:}{}

\Fn{\FOptimize{$\mathbf{s}, \mathbf{x}_s, \theta, \eta_\theta, \beta, B$}}{
    $\mathbf{o} \gets \mathcal{H}(\mathbf{s}\mid \mathbf{x}_s)$ \tcp*{compute partial observation}\
    $(\mathbf{a}, \log \pi_{\mathbf{a}}) \gets \mathrm{PC}\pi_\theta(\cdot \mid \mathbf{o}, \mathbf{x}_s)$ \tcp*{sample action, log-prob}\
    $\theta \leftarrow \theta - \eta_\theta \nabla_\theta \frac{1}{B}\!\sum_b [\beta\log \pi_{\mathbf{a}} - \min_j Q_{\phi_j}(\mathbf{s}, \mathbf{a})]$ \tcp*{update actor parameters}\
    \KwRet{$\theta$}\;
}
\end{algorithm}

\begin{algorithm}[H]
\caption{Critic Update Function}\label{alg:alphaopt}
\SetKwFunction{FOptimize}{UpdateQ}
\SetKwProg{Fn}{Function}{:}{}

\Fn{\FOptimize{$\mathbf{s}, \mathbf{a}, r, \mathbf{s}' , \mathbf{x}_s, \{\phi_j\}_{j=1}^{N}, \eta_\beta, \beta, B$}}{
    $\mathbf{o}' \gets \mathcal{H}(\mathbf{s}'\mid \mathbf{x}_s)$ \tcp*{compute partial observation}\
    $(\mathbf{a}', \log \pi_{\mathbf{a}'}) \gets \mathrm{PC}\pi_\theta(\cdot \mid \mathbf{o}', \mathbf{x}_s)$ \tcp*{sample action, log-prob}\
    $Q_{\text{target}} \leftarrow r + \gamma\!\left[\min_j Q_{\phi'_j}(\mathbf{s}', \mathbf{a}') - \beta \log \pi_{\mathbf{a}'} \right]$\;
    $\phi_j \leftarrow \phi_j - \eta_\phi \nabla_{\phi_j} \frac{1}{B}\!\sum_b (Q_{\phi_j}(\mathbf{s},\mathbf{a}) - Q_{\text{target}})^2,\; \forall j$ \tcp*{update critic ensemble}\
    \KwRet{$\{\phi_j\}_{j=1}^{N}$}\;
}
\end{algorithm}

\end{document}

%% file: method.tex
\section{Methodology}
\label{sec:Method}

\begin{figure}[H]
\centering
\includegraphics[width=0.9\linewidth]{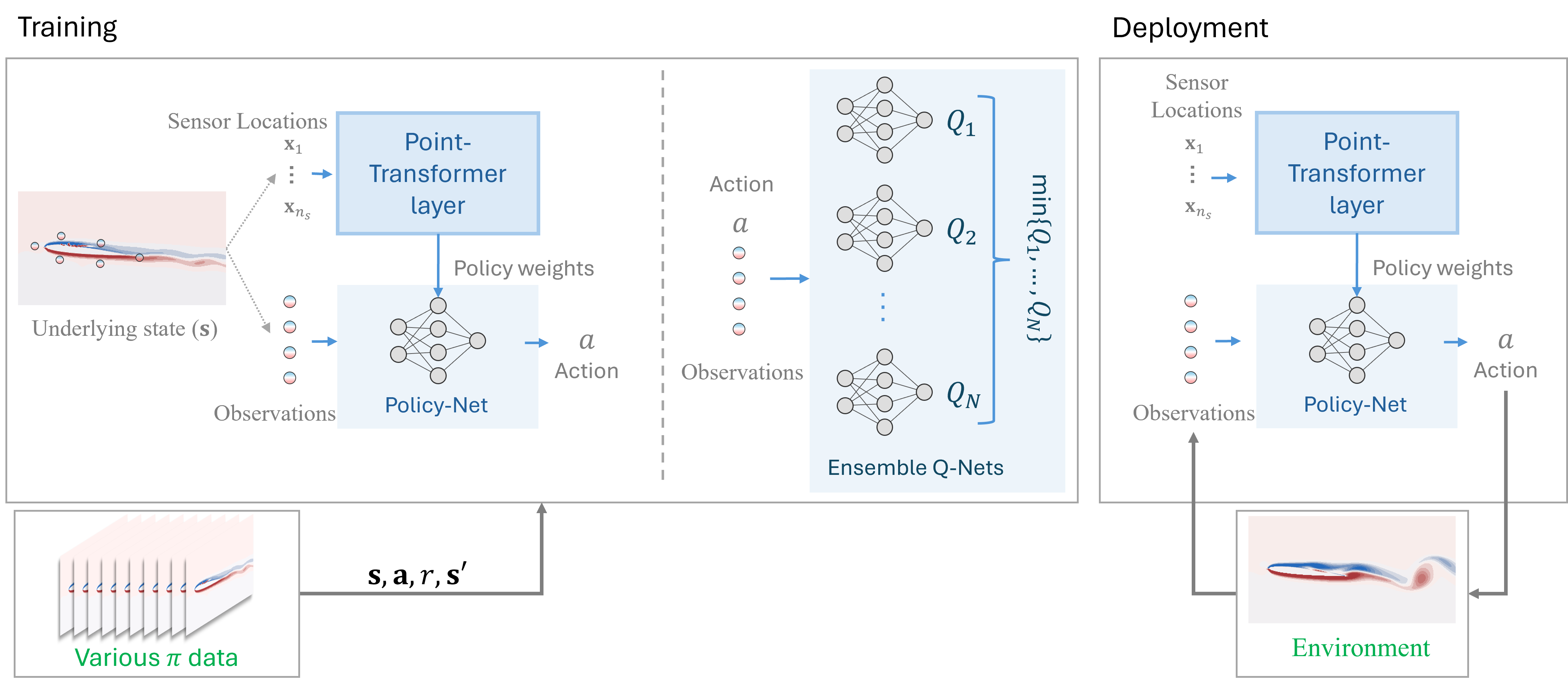}
\caption{Overall framework. \textbf{Dataset}: generate various $\pi$ data offline, with full states $\textbf{s}, \textbf{s}'$, actions $\mathbf{a}$ and and rewards $r$. \textbf{Training}: train offline with PC$\pi$-net: the actor net takes observations as input. The actor net policy weights is parameterzied by sensor locations through Point-Transformer layer using a hypernet-style weight generation. And the critic net is an ensemble with conservative target $\min_j Q_{\phi'_j}$ to mitigate distribution-shift issue. \textbf{Deployment}: deploy under a any sensor layout $\{\textbf{x}\}_{i=1}^{n_s}$ without retraining.}
\label{fig:offline_rl_workflow_clean}
\end{figure}

\subsection{Problem Formulation}
\label{subsec: problem}
Learning control policies for fluid systems to achieve objectives such as drag reduction has long been a challenging task. While reinforcement learning (RL) offers a promising approach, it typically relies on interactions with the environment, which can be costly, time-consuming, or even dangerous. For instance, allowing an aircraft to learn to fly through real-world trial-and-error or enabling a self-driving car to learn by directly operating on public roads could lead to catastrophic consequences.

In contrast, large volumes of data are continuously generated by human operators—for example, from piloted flights or road vehicles—which motivates the use of pre-collected data for policy learning. This paradigm, known as offline Reinforcement Learning (offline RL), seeks to learn effective policies entirely from offline datasets, without requiring additional environment interactions during training.

% \begin{figure}[!htp]
%     \centering
%     \includegraphics[width=\linewidth]{./fig/OfflineRL.png}
%     \caption{Illustration of the difference between Online and Offline RL, emphasizing the value of Offline RL for policy learning without environment interaction. \textcolor{red}{LS:Is this figure a must or need a refinement?}}
%     \label{fig:OfflineRL}
% \end{figure}

The focus of this work is to develop and evaluate offline RL methods for fluid dynamics applications. Since fluid environments are typically governed by partial differential equations (PDEs), we design two simulation environments based on governing fluid dynamics equations to serve as testbeds for our framework.

% \begin{figure}[!htp]
%     \centering
%     \includegraphics[width=\linewidth]{./fig/OfflineRL_policy_extract.png}
%     \caption{Overview of the proposed framework for extracting policies from data. Here, $\pi_\beta$ denotes the behavior policy used to generate the dataset--it may not be optimal or expert-generated--while $\pi_\theta$ represents the policy learned from that data. Multiple task-specific policies can be extracted from a single dataset using this approach.}
%     \label{fig:OfflineRL_pi_extract}
% \end{figure}

While learning a policy from data is valuable, it becomes inefficient when a new sensor configuration requires retraining. To address this limitation, we propose a sensor position-conditioned policy network (PC$\pi$-net) that can represent multiple policies corresponding to different sensor placements. In addition, we introduce a training strategy that enables the network to learn generalizable behavior across a range of observation configurations. 
% \todo[inline]{In my understanding, the $\mathbf{position-conditioned}$ are both in network structure and training strategy. In NN architecture, it uses point Transformer layers to encode different locations and in during it introduce perturbation and sampled from a distribution, is that right?}
This effectively defins a multi-observation framework and enables the the network to learn a control strategy robust to changes in sensor placement. The overall framework is shown in Fig.~\ref{fig:offline_rl_workflow_clean}

\subsection{Reinforcement Learning Overview}

Reinforcement Learning (RL) is a framework in which an agent interacts with an environment \(\mathcal{E}\) to learn a policy \(\pi(\mathbf{a}|\mathbf{s})\) that maps states \(\mathbf{s} \in \mathcal{S}\) to actions \(\mathbf{a} \in \mathcal{A}\), aiming to maximize cumulative rewards. Initially, the environment state \(\mathbf{s}_0\) is sampled from an initial state distribution \(d_0(\cdot)\), i.e., \(\mathbf{s}_0 \sim d_0(\cdot)\). At each subsequent control step \(t\), the agent observes the current state \(\mathbf{s}_t\), selects an action \(\mathbf{a}_t\) based on its policy \(\pi(\cdot|\mathbf{s}_t)\), receives a reward \(r(\mathbf{s}_t, \mathbf{a}_t): \mathcal{S}\times \mathcal{A} \rightarrow \mathbb{R}\), and transitions to a new state \(\mathbf{s}_{t+1}\) according to a transition probability \(T(\mathbf{s}_{t+1}|\mathbf{s}_t, \mathbf{a}_t)\) governed by the environment's dynamics.

The primary objective in RL is to maximize the expected cumulative discounted reward, defined as:
\[
J(\pi) = \mathbb{E}_{\pi} \left[ \sum_{t=0}^{\infty} \gamma^t r(\mathbf{s}_t, \mathbf{a}_t) \right],
\]
where \(\gamma \in [0,1)\) is the discount factor.
Formally, RL problems are typically modeled as Markov Decision Processes (MDPs), defined by the tuple \((\mathcal{S}, \mathcal{A}, T, r, d_0, \gamma)\).

% \subsubsection{Partially Observed Markov Decision Processes (POMDPs)}

In many real-world control tasks, particularly environments involving fluids, the agent does not have access to the full system state $\mathbf{s}_t$ at each time step. Instead, it receives a partial observation $\mathbf{o}_t \in \mathcal{O}$, which depends on the underlying state $\mathbf{s}_t$. Such problems are modelled as Partially Observed Markov Decision Processes (POMDPs), extending the standard MDP framework. A POMDP is defined by the tuple $(\mathcal{S}, \mathcal{A}, \mathcal{O}, T, E, r, d_0, \gamma)$, where $\mathcal{S}, \mathcal{A}, T, r, d_0$, and $\gamma$ are defined as before, $\mathcal{O}$ is a set of observations, where observation at control step $t$ is given by $\mathbf{o}_t \in \mathcal{O}$, and $E$ is an emission function, which defines the distribution $E(\mathbf{o}_t|\mathbf{s}_t)$. 
In a POMDP, the agent must base its policy $\pi(\mathbf{a}_t \mid \mathbf{o}_{\leq t})$ on the current and past observations, which may only partially capture the true state. In practice, augmenting observation with a fixed length of history ($o_t$, $o_{t-1}$, $o_{t-2}$ ...) is helpful on reducing the partial observability. In this work, however, we restrict the policy input to be only the current observation $o_t$ to maintain a low dimensional input space and to isolate the impact of sensor placement. The effect of temporal history integration is left for future work.

\subsection{Offline Reinforcement Learning}
In traditional RL, also called online RL, settings, the agent learns through continuous interaction with the environment, collecting new data to improve its policy iteratively. However, in many real-world scenarios, such interactions are expensive, time-consuming, or unsafe, making online training impractical. Offline RL addresses this issue by enabling policy learning directly from pre-collected datasets, which is the focus of our work.

Offline RL circumvents the need for active environment interaction by decoupling policy learning from data collection. Instead of exploring in real time, the agent learns from a static dataset collected by a previously deployed policy—such as a human expert, a heuristic controller, or an earlier learned policy. This setup eliminates the risks and costs associated with online exploration. However, it introduces new challenges, such as distributional shift and extrapolation error, where the learned policy may query out-of-distribution actions not supported by the dataset. To address this, modern offline RL algorithms incorporate techniques like policy regularization, conservative Q-function estimation, or ensemble learning to ensure that the learned policy remains close to the data distribution and avoids unreliable predictions in underrepresented regions of the state-action space. 
% LS: A key challenge in offline RL is the distributional shifts, which means the learned policy may query the state-action pairs that are out of the training data distributions. In fluid control tasks, such actions are not reliable and can correspond to actuator commands that are far from the stable operating conditions, potentially degrading the control performances. 
% In the current work, we adopt an SACN framework with an ensemble-based conservative critic and provide pessimistic estimation for poorly supported actions, therefore promoting the policies based on actions well-represented in the training dataset.

\subsection{Offline SACN} 
\label{subsec: offline_SAC_N}
% \todo[inline]{In my understand Section 2.1-2.4 are background and section 2.5 is the novelty in this method?}

Q-learning~\cite{mnih2015human, haarnoja2018soft} is a widely adopted approach for learning control policies by estimating a state-action value function \( Q_\phi(\mathbf{s}, \mathbf{a}) \), which represents the expected cumulative reward obtained by taking action \( \mathbf{a} \) in state \( \mathbf{s} \) and subsequently following a given policy. In actor–critic frameworks, this Q-function is parameterized by a neural network, called Q-network, and trained alongside the policy by minimizing the Bellman residual:
$(Q_\phi(\mathbf{s}, \mathbf{a})-\mathcal{B}^{\pi_\theta}Q_\phi(\mathbf{s}, \mathbf{a}))^2$, where $\mathcal{B}^{\pi_\theta}$ is the Bellman operator given as
$$\mathcal{B}^{\pi_\theta}Q_\phi(\mathbf{s}, \mathbf{a})=\mathbb{E}_{\mathbf{s}'\sim T(\cdot|\mathbf{s}, \mathbf{a})}\Big[ r(\mathbf{s}, \mathbf{a}) + \gamma \mathbb{E}_{\mathbf{a}'\sim \pi_\theta (\cdot|\mathbf{s}')}[Q_{\phi}(\mathbf{s}', \mathbf{a}')] \Big]$$

We simplify the notation for current state $\mathbf{s}_t$, action $\mathbf{a}_t$, and next state $\mathbf{s}_{t+1}$ as $\mathbf{s}$, $\mathbf{a}$, and $\mathbf{s}'$, respectively.
Soft Actor-Critic (SAC)~\cite{haarnoja2018soft} is a state-of-the-art actor–critic algorithm that incorporates entropy regularization into the policy objective to promote exploration and stability. To address the overestimation bias often encountered in value estimation, SAC utilizes Clipped Double Q-learning~\cite{fujimoto2018addressing}, in which two independent Q-networks are trained, and the minimum of their outputs is used to form the Bellman target:
$y := r(\mathbf{s}, \mathbf{a}) + \gamma \, \mathbb{E}_{\mathbf{a}' \sim \pi_\theta(\cdot|\mathbf{s}')} \left[ \min_{j=1,2} Q_{\phi'_j}(\mathbf{s}', \mathbf{a}') \right].$

SACN~\cite{an2021uncertainty} builds upon this idea by generalizing from two to an ensemble of \( N \) Q-networks. Each Q-network in the ensemble independently estimates the value function, and the Bellman target is computed using the minimum value across all members. This ensemble-based approach provides a more conservative and robust estimation of Q-values, which is especially critical in the offline RL setting where out-of-distribution (OOD) actions—those not well represented in the dataset—can lead to severe performance degradation.

From a probabilistic viewpoint, each Q-network can be interpreted as a sample from the underlying epistemic (model) uncertainty in approximating the true value function. With only two networks, uncertainty estimates are limited and potentially unreliable. By increasing the ensemble size, SACN captures this uncertainty more accurately, leading to more informed and cautious value estimates. This helps the learned policy avoid unsafe regions of the action space and stay close to the support of the dataset, ultimately enabling more stable and reliable policy extraction in offline scenarios.

In SACN, the objective for each Q-network is to minimise $J_q(Q_{\phi_i})$ defined as
\begin{equation}
    J_q(Q_{\phi_i}) \coloneqq \mathbb{E}_{(\textbf{s},\textbf{a},\textbf{s}',r) \sim \mathcal{D}} \Bigg[ \Bigg( Q_{\phi_i}(\textbf{s},\textbf{a}) - \bigg(r+\gamma \mathbb{E}_{\textbf{a}'\sim \pi_{\theta}(\cdot|\textbf{s}')} \Big[ \min_{j=1,\cdots,N} Q_{\phi'_j}(\textbf{s}', \textbf{a}') - \beta \log\pi_\theta(\textbf{a}'|\textbf{s}') \Big] \bigg) \Bigg)^2 \Bigg].
\end{equation}
% \begin{subequations}
% \begin{align}
%     &J_q(Q_{\phi_i}) \coloneqq \mathbb{E}_{(\textbf{s},\textbf{a},\textbf{s}',r) \sim \mathcal{D}} \Bigg[ \bigg( Q_{\phi_i}(\textbf{s},\textbf{a}) - \big(r+\gamma \mathbb{E}_{\textbf{a}'\sim \pi_{\theta}(\cdot|\textbf{s}')} [ Q_{\phi'}(\textbf{s}', \textbf{a}') ] \big) \bigg)^2 \Bigg], \\
%     &Q_{\phi'}(\textbf{s}', \textbf{a}') = \min_{j=1,\cdots,N} Q_{\phi'_j}(\textbf{s}', \textbf{a}') - \beta \log\pi_\theta(\textbf{a}'|\textbf{s}').
% \end{align}
% \end{subequations}
Here $Q_{\phi'_j}$ represents the target Q-networks which are softly updated to ensure smooth and stable learning. Specifically, the parameters \( \phi'_j \) of the target Q-networks are updated as a moving average of the Q-network parameters \( \phi_j \) using:
$\phi'_j \leftarrow \tau \phi_j + (1 - \tau) \phi'_j,$
where \( \tau \in (0, 1) \) is the soft update coefficient, typically set to a small value (e.g., 0.005). This gradual update reduces oscillations and divergence during training.

The policy, which is also parameterised by a neural network, is updated in an alternating fashion to maximise the $J_p(\pi_\theta)$ defined as
\begin{equation}
    J_p(\pi_\theta) := \mathbb{E}_{\textbf{s}\sim\mathcal{D}, \textbf{a}\sim\pi_\theta(\cdot|\textbf{s})}\Bigg[ \min_{j=1,\cdots,N} Q_{\phi_j}(\textbf{s}, \textbf{a}) - \beta \log\pi_\theta(\textbf{a}|\textbf{s}) \Bigg].
\end{equation}

For POMDP, the above equation is only modified such that $\pi_\theta(\textbf{a}|\textbf{s})$ is replaced with $\pi_\theta(\textbf{a}|\textbf{o})$. If the data contains the full state, we have the choice to use the full state to train the Q-network. 
For current study we focus on local information, so the observation $\mathbf{o}$ is deterministically generated from the full state $\mathbf{s}$ by an observation operator $\mathcal{H}(\cdot|\mathbf{x}_s)$, i.e. $\mathbf{o}=\mathcal{H}(\mathbf{s}|\mathbf{x}_s)$, which samples the state at a set of fixed sensor locations \( \mathbf{x}_s = \{\mathbf{x}_i\}_{i=1}^{n_s}, \mathbf{x}_i \in \mathbb{R}^{3}\). Accordingly, the emission function is given as $E(\mathbf{o}|\mathbf{s})=\delta(\mathbf{o}-\mathcal{H}(\mathbf{s}|\mathbf{x}_s))$ and the SACN algorithm for POMDP becomes
\begin{subequations}
\begin{align}
    &J_q(Q_{\phi_i}) \coloneqq \mathbb{E}_{(\textbf{s},\textbf{a},\textbf{s}',r) \sim \mathcal{D}} \Bigg[ \Bigg( Q_{\phi_i}(\textbf{s},\textbf{a}) - \bigg(r+\gamma \mathbb{E}_{\textbf{a}'\sim \pi_{\theta}(\cdot|\textbf{o}')} \Big[ \min_{j=1,\cdots,N} Q_{\phi'_j}(\textbf{s}', \textbf{a}') - \beta \log\pi_\theta(\textbf{a}'|\mathbf{o}') \Big] \bigg) \Bigg)^2 \Bigg], \\
    &J_p(\pi_\theta) := \mathbb{E}_{\textbf{s}\sim\mathcal{D}, \textbf{a}\sim\pi_\theta(\cdot|\mathbf{o})}\Bigg[ \min_{j=1,\cdots,N} Q_{\phi_j}(\textbf{s}, \textbf{a}) - \beta \log\pi_\theta(\textbf{a}|\mathbf{o}) \Bigg].
\end{align}
\end{subequations}
% \todo[inline]{LS: it is harder to digest and seems missing a transition between POMDP and SACN settings. Need double check}
The standard POMDP policy is trained with a fixed set of sensor locations $\mathbf{x}_s$. %However, it is sampled from a distribution during training in our multi-observation framework and could vary during deployment. This modification facilitates the policy to generalize over different sensor configurations.

% For our current study, we assume the data will contain the full state, however we intend to learn a policy that works on a few sensors represented by partial observation $o$.

% \subsection{Sensor Aware Policy Network (PC$\pi$-net)}
\subsection{Sensor Position-Condition Policy Network (PC$\pi$-net)}
\label{subsec:SApi}
In settings where full state information is available in the offline dataset, it becomes possible to train policies that operate under various partial observation configurations. This opens up opportunities for systematically exploring and optimizing sensor placements to identify configurations that yield the best policy performance. However, using SACN for this purpose typically requires training a separate policy network for each observation setup, which can be computationally expensive and time-consuming.

To address this limitation, we propose a sensor position-conditioned policy network (PC$\pi$-net) that enables a single policy model to generalize across multiple observation configurations. Instead of training distinct networks for each sensor layout, our approach conditions the policy on the sensor configuration provided at deployment. As a result, the same trained network can generate appropriate actions for a variety of sensor setups.
This design eliminates the need for redundant training runs and allows for rapid evaluation of different observation strategies within a single deployment. Consequently, it significantly accelerates the process of identifying optimal sensor placements for high-performing policies, making it especially valuable in experimental or resource-constrained environments.

% \todo[inline]{Anything useful needs to have the invaraince properties in our application like permutation invariance, that's good to highlight.}
\begin{figure}[!ht]
    \centering
    \includegraphics[width=\linewidth]{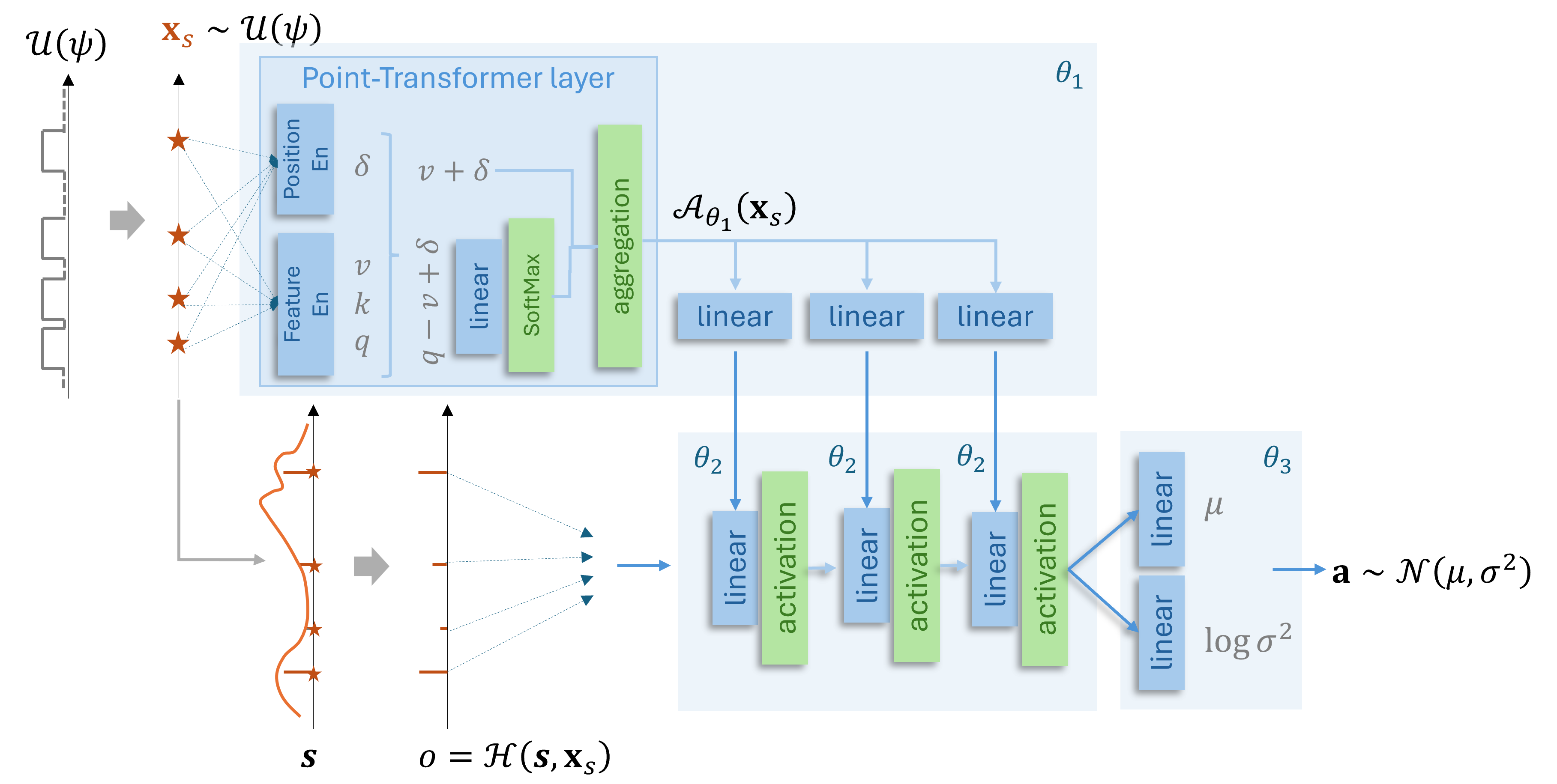}
    \caption{Architecture of sensor position-conditioned policy network (PC$\pi$-net). Sensor positions $\textbf{x}_s$, sampled from $\mathcal{U}(\psi)$, are used as conditioning input and processed through a point-transformer module to generate embeddings. These embeddings are then upscaled to parameterize the weights of the policy network, which maps observations $\textbf{o}$ to an action distribution $\mathcal{N}(\mu, \sigma^2)$.}
    \label{fig:MultiPolicy_Net}
\end{figure}

We define a position-conditioned optimal policy $PC\pi(\textbf{a}|\mathbf{o}, \textbf{x}_s)$
% $\pi(\textbf{a}|\mathcal{H}(\mathbf{s}|\mathbf{x}_s), \textbf{x}_s)$
that maximizes the same cumulative discounted reward $\mathbb{E}_{\pi}[\sum_{t=0}^{\infty} \gamma^t r(\textbf{s}_t, \textbf{a}_t)]$ for given set of sensor positions \( \mathbf{x}_s\). Here $\mathbf{o}=\mathcal{H}(\mathbf{s}|\mathbf{x}_s)$. To enable flexible decision-making across different sensor placements, we design the PC$\pi$-Net architecture as illustrated in Fig.~\ref{fig:MultiPolicy_Net}. The architecture consists of a hypernetwork and a policy network: the hypernetwork incorporates a Point Transformer to encode sensor positions $\textbf{x}_s$ and outputs the parameters (weights and biases) of the policy network. The policy network, implemented as an MLP, maps the observation $\textbf{o}$ to an action distribution $PC\pi_{\theta}(\textbf{a}|\mathbf{o}, \textbf{x}_s)$ conditioned on the sensor layout. Formally,
\begin{subequations}
\begin{alignat}{3}
    \theta_2(\textbf{x}_s) & = l^p_{\theta_1} \big( \mathcal{A}_{\theta_1}(\textbf{x}_s)\big),  \\
    y &= NN_{\theta_2(\textbf{x}_s)}(\mathcal{H}(\mathbf{s}|\mathbf{x}_s)), \\
    PC\pi_{\theta}(\textbf{a}|\mathbf{o}, \textbf{x}_s) &= \mathcal{N}\big(l^{\mu}_{\theta_3}(y), \exp(l^{\sigma}_{\theta_3}(y))\big),
\end{alignat}
\end{subequations}
where $\theta = \{\theta_1, \theta_2, \theta_3\}$ are trainable parameters. The module \( \mathcal{A}_{\theta_1}(\cdot) \) denotes the Point Transformer layer~\cite{zhao2021point}, ($l^p,l^{\mu},l^{\sigma}$) represents the dedicated linear layers, and \( NN(\cdot) = l\cdot\text{ReLU}\cdot l\cdot\text{ReLU}\cdot l \cdot \text{ReLU}(\cdot)\) is a three-layer feedforward NN. A schematic of this architecture is illustrated in Fig.~\ref{fig:MultiPolicy_Net}.

\subsubsection*{Point Transformer Layer}

To effectively encode spatial relationships between sensors, we employ a Point Transformer layer based on the vector self-attention framework of~\cite{zhao2021point}. Given a set of sensor positions \( \mathbf{x}_s \in \mathbb{R}^{n_s \times 3} \), the Point Transformer performs self-attention over the sensor points, incorporating both geometric proximity and learned features. The core mechanism is defined as~
% \todo[inline]{are you sure it is $q(e_i)-k(e_j)^{T}$? Commonly, it is $q\cdot k$ in attention. Is the actual encoding $\delta_{ij}$ actually added to $v$ or concatenated to it?}:
\begin{subequations}
\begin{alignat}{4}
    \textbf{e}_i &= \mathcal{E}(\textbf{x}_i), \\
    \delta_{ij} &= l(\textbf{x}_i - \textbf{x}_j),  \\
    a_i &= \sum_{\textbf{x}_j \in \textbf{x}_s} \text{Softmax}\Big(\rho\big[q(\textbf{e}_i)-k(\textbf{e}_j)^T+\delta_{ij}\big]\Big) \odot(v(\textbf{e}_j)+\delta_{ij}),\\
    \mathcal{A}_{\theta_1}(\textbf{x}_s) &=  \frac{1}{n}\sum_{i=1}^n a_i
\end{alignat}
\end{subequations}
% \begin{subequations}
% \begin{alignat}{3}
%     \textbf{e}_s &= \mathcal{E}_{\theta_1}(\textbf{x}_s), \\
%     [\delta] &= l(\textbf{x}_s - \textbf{x}_s^T),  \\
%     P_{\theta_1}(\textbf{x}_s) &= \sum_{\textbf{x}_s} \text{Softmax}\big(A(q(\textbf{x}_s)-k(\textbf{x}_s)^T+[\delta])\big) \odot(v(\textbf{x}_s)+[\delta]),
% \end{alignat}
% \end{subequations}
where \( q(\cdot), k(\cdot), v(\cdot) \) are pointwise feature transformations, linear projections in this study; \( \delta_{ij} \) encodes relative positional information using a learned Multi-Layer Perceptron (MLP); \( \mathcal{E}(\cdot) = l \cdot\text{LayerNorm}\cdot\text{ReLU} \cdot l \cdot\text{LayerNorm}\cdot\text{ReLU} \cdot l(\cdot) \) is a three-layer NN and \( \rho(\cdot) = l\cdot\text{ReLU}\cdot l(\cdot) \) is a two-layer NN. The attention \( a_i \) integrates contextual information across all sensor positions, weighted by relevance and spatial structure.

This design enables the policy network to adaptively condition its behavior on varying sensor configurations, capturing not only individual sensor features but also their geometric relationships, which is critical for generalization across sensor placements. An important property of the Point Transformer is its permutation invariance with respect to input points, meaning that the output remains consistent regardless of the ordering of sensor positions. This is particularly beneficial in sensor configuration tasks, where there is no inherent order among sensor locations. Permutation invariance allows the model to focus on spatial relationships and sensor interactions rather than being biased by arbitrary indexing, thereby enhancing robustness and transferability across different observation layouts.

\subsubsection*{Position-Conditioned SACN (PC-SACN)} 
% \begin{figure}[H]
% \resizebox{\textwidth}{!}{%
% \begin{minipage}{\textwidth}
% \begin{algorithm}[H]
% \caption{position-conditioned SACN (PC-SACN)}
% \label{alg:PC-SACN}
% \KwIn{$D$: offline dataset; $\psi$: mean sensor layout; $\Delta$: perturbation; $N$: ensemble size; $\gamma$: discount; $\tau$: target rate; $\beta$: temperature; $\eta_Q,\eta_\pi,\eta_\psi$: learning rates; $B$: batch size}
% \KwOut{Actor $\mathrm{SA}\pi_\theta$, critics $\{Q_{\phi_j}\}$}
% Initialize $\theta, \{\phi_j\}_{j=1}^N, \{\phi'_j \!\leftarrow\! \phi_j\}$ and $\psi$\;
% \Repeat{convergence}{
%     Sample $\{(s,a,r,s')\}_{b=1}^B \!\sim\! D$\;
%     Sample $x_s^{(b)} \!\sim\! \mathcal{U}_\psi$, compute $o_b = H(s_b|x_s^{(b)}),\; o'_b = H(s'_b|x_s^{(b)})$\;
%     Sample $a'_b,\log\pi'_b \!\sim\! \mathrm{SA}\pi_\theta(\cdot|o'_b,x_s^{(b)})$\;
%     $y_b \leftarrow r_b + \gamma\!\left[\min_j Q_{\phi'_j}(s'_b,a'_b) - \beta \log\pi'_b\right]$\;
%     $\phi_j \leftarrow \phi_j - \eta_Q \nabla_{\phi_j} \frac{1}{B}\!\sum_b (Q_{\phi_j}(s_b,a_b) - y_b)^2,\; \forall j$\;
%     Sample $a^\theta_b,\log\pi_b \!\sim\! \mathrm{SA}\pi_\theta(\cdot|o_b,x_s^{(b)})$\;
%     $\theta \leftarrow \theta - \eta_\pi \nabla_\theta \frac{1}{B}\!\sum_b [\beta\log\pi_b - \min_j Q_{\phi_j}(s_b,a^\theta_b)]$\;
%     \If{optimize sensors}{
%         $\psi \leftarrow \psi + \eta_\psi \nabla_\psi \frac{1}{B}\!\sum_b \min_j Q_{\phi'_j}(s_b,a_b^{\psi})$\;
%     }
%     $\phi'_j \leftarrow \tau \phi_j + (1-\tau)\phi'_j,\; \forall j$\;
% }
% \end{algorithm}
% \end{minipage}
% }
% \end{figure}

To enable a unified position-conditioned policy $\pi(\textbf{a}|\textbf{o}, \textbf{x}_s)$, we replace the MLP actor network in SACN with the proposed PC$\pi$-net. This network is trained using observations $o = \mathcal{H}(\mathbf{s}|\mathbf{x}_s)$ derived from full state information \( \mathbf{s} \).

The actor and critic objectives are modified to account for the varying sensor configurations as follows:
\begin{subequations}\label{eq:PC-SACN-opt}
\begin{alignat}{2}
    & \min_{\phi_i} \quad \mathbb{E}_{\textbf{x}_s\sim \mathcal{U}_{\psi}, (\textbf{s},\textbf{a},\textbf{s}', r) \sim \mathcal{D}} \Bigg[ \Bigg( Q_{\phi_i}(\textbf{s},\textbf{a}) - \Bigg(r+\gamma \mathbb{E}_{\textbf{a}'\sim PC\pi_{\theta}(\cdot|\mathcal{H}(\mathbf{s}'|\mathbf{x}_s),\mathbf{x}_s)} \bigg[ \min_{j=1,\cdots,N} Q_{\phi'_j}(\textbf{s}', \textbf{a}') 
    & \notag \\
    & \hspace{8.5cm}- \beta \log \Big[PC\pi_\theta(\textbf{a}'|\mathcal{H}(\mathbf{s}'|\mathbf{x}_s),\mathbf{x}_s)\Big] \bigg] \Bigg) \Bigg)^2 \Bigg],  \\
    & \max_{\theta} \quad \mathbb{E}_{\textbf{x}_s\sim \mathcal{U}_{\psi},\textbf{s}\sim\mathcal{D}, \textbf{a}\sim PC\pi_\theta(\cdot|\mathcal{H}(\mathbf{s}|\mathbf{x}_s),\mathbf{x}_s)} \bigg[ \min_{j=1,\cdots,N} Q_{\phi_j}(\textbf{s}, \textbf{a}) - \beta \log \Big[PC\pi_\theta(\textbf{a}|\mathcal{H}(\mathbf{s}|\mathbf{x}_s),\mathbf{x}_s)\Big] \bigg].
\end{alignat}
% \todo[inline]{May need a clearer way to distinguish if it is log(SA) or logS, adding another bracket? though it may looks very ugly.}
\end{subequations}
The temperature parameter $\beta$ plays a critical role in balancing reward maximization and exploration. A higher $\beta$ encourages stochasticity in the policy by weighting the entropy term more strongly, while a lower $\beta$ drives the policy toward determinism. Instead of treating $\beta$ as a fixed hyperparameter, we optimise it automatically to maintain a target entropy level $H_{\text{target}}$, ensuring adaptive control over exploration throughout training. This is achieved as
\begin{equation}\label{eq:beta-opt}
    \min_{\beta} \quad \mathbb{E}_{\textbf{x}_s\sim \mathcal{U}_{\psi},\textbf{s}\sim\mathcal{D}, \textbf{a}\sim PC\pi_\theta(\cdot|\mathcal{H}(\mathbf{s}|\mathbf{x}_s),\mathbf{x}_s)} -\log\beta \Big(\log \big[PC\pi_\theta(\textbf{a}|\mathcal{H}(\mathbf{s}|\mathbf{x}_s),\mathbf{x}_s)\big] + H_{\text{target}}\Big).
\end{equation}
To train across a diverse range of sensor configurations, we define a perturbation range $\Delta$ around each sensor's nominal position \( \psi_i \in \mathbb{R}^d \). At each training step, the sensor positions \( \mathbf{x}_s = \{\mathbf{x}_i\}_{i=1}^{n_s} \) are sampled from a uniform distribution centered around \( \psi_i \), i.e.,
$ \textbf{x}_i \sim \psi_i + (2\mathcal{U}(\cdot)-1)\Delta$
where \( \mathcal{U}(\cdot) \) denotes sampling from the unit uniform distribution.
% \todo[inline]{is this \textbf{2U-1} rangeing from 0 to 1?, why not just use \textbf{U}? Is it \textbf{unit uniform distribution} or \textbf{uniform distribution}}.
This strategy enables the network to generalize across small variations in sensor placement and learn robust policies adaptable to different observation configurations.

Additionally, during training, the sensor placement can be optimised; we introduce an auxiliary objective to optimize the sensor placement parameters \( \psi = \{\psi_i\}_{i=1}^{n_s} \) themselves. This objective aims to identify sensor configurations that maximize expected value under the learned policy:
\begin{equation}
    \max_{\psi} \quad \mathbb{E}_{\textbf{x}_s\sim \mathcal{U}_{\psi},\textbf{s}\sim\mathcal{D}, \textbf{a}\sim PC\pi_\theta(\cdot|\mathcal{H}(\mathbf{s}|\mathbf{x}_s),\mathbf{x}_s)}\Bigg[ \min_{j=1,\cdots,N} Q_{\phi'_j}(\textbf{s}, \textbf{a}) - \beta \log \Big[PC\pi_\theta(\textbf{a}|\mathcal{H}(\mathbf{s}|\mathbf{x}_s),\mathbf{x}_s)\Big]\Bigg]
    \label{eq:sen_opt_obj}
\end{equation}
% \todo[inline]{How do you actually optimize objective function 7? It is not clear here. Is it an additional loss term?}
Eq.~\ref{eq:PC-SACN-opt}a, Eq.~\ref{eq:PC-SACN-opt}b, Eq.~\ref{eq:beta-opt}, and  Eq.~\ref{eq:sen_opt_obj} are optimized sequentially at every epoch. This joint training framework allows us not only to learn a generalized position-conditioned policy but also to identify optimal sensor layouts for improved decision performance. Detailed algorithm can be found in \ref{sec:PC-SACN-algo}.

% \begin{algorithm}[H]
% \caption{position-conditioned SACN (PC-SACN)}
% \label{alg:PC-SACN}
% \KwIn{$D$: offline dataset; $\psi$: mean sensor layout; $\Delta$: perturbation; $N$: ensemble size; $\gamma$: discount; $\tau$: target rate; $\beta$: temperature; $\eta_Q,\eta_\pi,\eta_\psi$: learning rates; $B$: batch size}
% \KwOut{Actor $\mathrm{SA}\pi_\theta$, critics $\{Q_{\phi_j}\}$}

% Initialize $\theta, \{\phi_j\}_{j=1}^N, \{\phi'_j \!\leftarrow\! \phi_j\}$ and $\psi$\; % initialize networks and sensor layout

% \For{each training step}{
%     $\{(\mathbf{s},\mathbf{a},r,\mathbf{s}')\}_{b=1}^B \!\sim\! \mathcal{D}$ \tcp*{sample a batch of data from the dataset}
%     $\mathbf{x}_s \gets \mathcal{U}_\psi$ \tcp*{sample sensor positions around $\psi$}
    
%     \If{optimize sensors}{
%         $\psi \leftarrow$ \text{OptimizeSensors}($\mathbf{s}, \mathbf{x}_s, \psi, \eta_\psi, \beta, B$) \tcp*{update mean sensor positions}
%     }
    
%     $\beta \leftarrow$ \text{UpdateBeta}($\mathbf{s}, \mathbf{x}_s, \eta_\beta, \beta, B$) \tcp*{adjust entropy temperature}
    
%     $\theta \leftarrow$ \text{UpdateActor}($\mathbf{s}, \mathbf{x}_s, \theta, \eta_\theta, \beta, B$) \tcp*{update actor parameters}
    
%     $\{\phi_j\}_{j=1}^{N} \leftarrow$ \text{UpdateQ}($\mathbf{s}, \mathbf{a}, r, \mathbf{s}' , \mathbf{x}_s, \{\phi_j\}_{j=1}^{N}, \eta_\beta, \beta, B$) \tcp*{update critic ensemble}
    
%     $\phi'_j \leftarrow \tau \phi_j + (1-\tau)\phi'_j,\; \forall j$ \tcp*{soft update of target critics}
% }
% \end{algorithm}

\subsection{Test Environments}
\label{subsec:test_env}
\subsubsection{Kuramoto–Sivashinsky (KS)}
The first control scenario is governed by one-dimensional partial differential equation (PDE) - the Kuramoto–Sivashinsky (KS) equation, which exhibits spatiotemprally chaotic or weakly turbulent behavior, and is widely used as a system for turbulence study~\cite{cvitanovic2010state}. In this case, the KS environment is controlled by four actuators distributed equally in space to minimize the energy dissipation and total input power. The physics of this system is governed by the KS equation,
\begin{equation}
    \frac{\partial u}{\partial t} + \frac{\partial^2 u}{\partial x^2} + \frac{\partial^4 u}{\partial x^4} + \frac{1}{2} u\frac{\partial u}{\partial x} = f(x,t), \ \    x \in [0,l], \ t\in[0,\infty),
    \label{eq:KS}
\end{equation}
where $u$ is the state variable, and $f$ represents the source term (i.e., actuator) defined by,
\begin{equation}
    f(x,t) = \sum_{i=1}^{4}\frac{a_i(t) e^{-(x-x_i)^2/2}}{\sqrt{2\pi}},
\end{equation}
where $x_i \in \{l/8,\,3l/8,\,5l/8,\,7l/8\}$ is the spatial locations of the actuator, and $\boldsymbol{a} = \{a_i(t)\}$, $ i=1,2,3,4 $, $a_i(t)\in [-0.5,0.5]$ defines the control parameters. 
To achieve the control goal, the reward function is defined as follows,
\begin{equation}
    r = -\frac{1}{T l}\int_{t_0}^{t_0+T}\int_0^l \left((\frac{\partial^2 u}{\partial x^2})^2 +(\frac{\partial u}{\partial x})^2 + u f\right)\,dx\, dt 
\end{equation}

\begin{figure}
    \centering
    \includegraphics[width=0.8\linewidth]{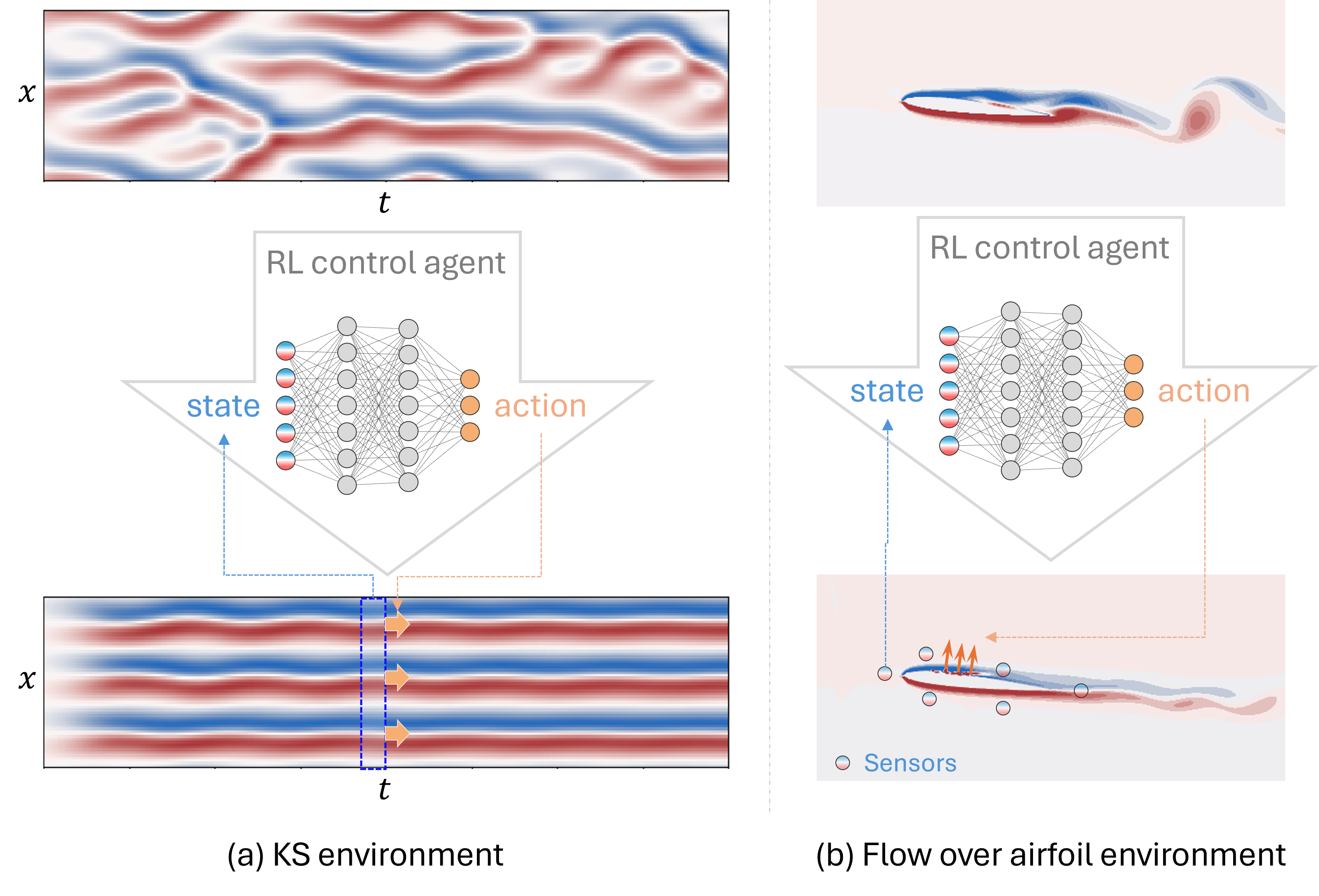}
    \caption{Schematic of control in (a) KS and (b) flow over airfoil environments. The top panels illustrate the system state without control, while the bottom panels show the corresponding state after control has been applied.}
    \label{fig:ks_af_env_schematic}
\end{figure}

\subsubsection{Flow over airfoil}
\begin{figure}[H]
    \centering
    \includegraphics[width=0.5\linewidth]{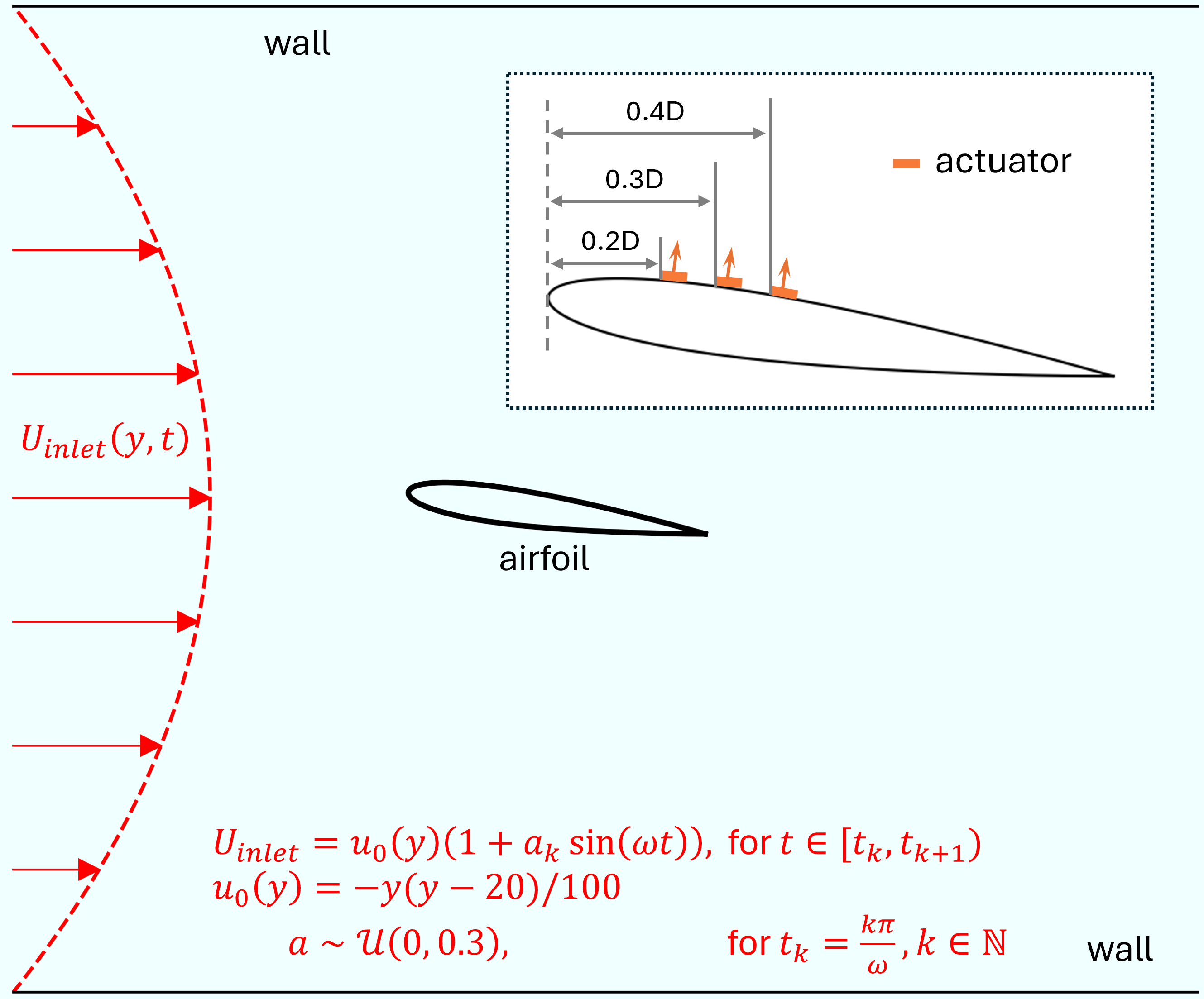}
    \caption{Schematic of the flow over airfoil environment. A fluctuating inlet velocity is applied to induce variations in lift and drag experienced by the airfoil. Three actuators are positioned along the upper surface of the airfoil to control the flow and regulate these fluctuations.}
    \label{fig:airfoil_env_schematic}
\end{figure}

The second control scenario is governed by a two-dimensional PDE - the Navier-Stokes (NS) equation, which simulate the incompressible-viscous flow given by
\begin{equation}
\label{eq:ns}
\begin{aligned}
\nabla\cdot\mathbf{u} = 0, \hspace{7em}&\mathbf{x}, t \in \Omega_f \times [0, T]\\
\frac{\partial{\mathbf{u}}}{\partial t} = -(\mathbf{u}\cdot\nabla){\mathbf{u}}+\nu\nabla^2\mathbf{u}-\frac{1}{\rho}\nabla{p}+\mathbf{f}_{s}+\mathbf{f}_{c}, \hspace{3em} &\mathbf{x}, t \in \Omega_f \times [0, T]
\end{aligned} 
\end{equation}
where $t$ and $\mathbf{x}$ denote the temporal and spatial coordinates in the Eulerian frame. The fluid velocity $\mathbf{u}(\mathbf{x}, t)$ and pressure $p(\mathbf{x}, t)$ are defined over the fluid domain $\Omega_f \subset \mathbb{R}^2$. The parameters $\rho$ and $\nu$ represent the fluid density and kinematic viscosity, respectively. The external body force $\mathbf{f}_s$ accounts for interactions with immersed boundaries, and $\mathbf{f}_c$ accounts for body force due to the control actuators. Additional details can be found in \ref{sec:airfoil_env}. 
For controlling flow over the airfoil, we used jets on top of the airfoil (similar to~\cite{wang2022deep}), as shown in Fig.~\ref{fig:airfoil_env_schematic}, which are emulated in the environment by body force $\mathbf{f}_c$ defined by
\begin{equation}
    \mathbf{f}_c = \sum_{i=1}^3 A_i(t)  \delta(x-x_{s_i}),
\end{equation}
where $x_{s_i} \in \{ [0.2D, 0.2D+0.2], [0.3D, 0.3D+0.2], [0.4D, 0.4D+0.2] \}$ is the spatial locations of the actuator as shown by orange color in Fig.~\ref{fig:airfoil_env_schematic}. The coefficients $A_i(t)$ are defined such that $\sum_{i=1}^3 A_i = 0$ and are defined by two control parameters $\{a_i\}_{i=1}^2$ as
\( A_1 = 100a_1, A_2 = -100(a_1+a_2), A_3 = 100a_2. \) 
% More details can be found in \ref{sec:airfoil_env}.

To introduce fluctuation in lift, an inlet flow as a function of time is given as
\begin{subequations}
    \begin{alignat}{3}
        U_{\text{inlet}}(y, t) &= u_0(y) \left(1 + a_k \sin(\omega t)\right), \quad \text{for } t \in \left[t_k, t_{k+1}\right), \\
        u_0(y) &= - y(y-20) / 100, \\
        a(t_k) &\sim \mathcal{U}(0, 0.3), \quad \text{for } t_k = k \cdot \frac{\pi}{\omega}, \quad k \in \mathbb{N},
    \end{alignat}
\end{subequations}
The inlet flow oscillates with frequency $\omega$ and the amplitude $a_k$ randomly varies after every time period of $\frac{\pi}{\omega}$.
The flow over the airfoil is controlled by three actuators distributed equally in space, as shown in Fig.~\ref{fig:airfoil_env_schematic}, to minimise the fluctuation in lift experienced by the airfoil and reduce drag. To achieve the control goal, the reward function is defined as follows,
\begin{equation}
    r = -|\bar{C_L}-m_1|-0.1|\bar{C_D}|
\end{equation}
where the lift and drag coefficients are defined as:
\begin{equation}
\begin{aligned}
C_L &= \frac{\int_{\Omega} \mathbf{F} \cdot \mathbf{n}_y \, \mathrm{d}\Omega}{0.5\, \rho\, \bar{U}_{\mathrm{inlet}}^2\, c} \\
C_D &= \frac{\int_{\Omega} \mathbf{F} \cdot \mathbf{n}_x \, \mathrm{d}\Omega}{0.5\, \rho\, \bar{U}_{\mathrm{inlet}}^2\, c}
\end{aligned}
\end{equation}
where $\mathbf{F}$ is the fluid force exerted on the airfoil, including both pressure and viscous contributions; $\rho$ is the fluid density; $c$ is the airfoil arc length; and $\mathbf{n}_x$ and $\mathbf{n}_y$ are the unit vectors in the streamwise $x$ and vertical $y$ directions, respectively.

%% file: result.tex
\section{Results}
\label{sec:Result}
We evaluate the performance of the proposed framework on two benchmark environments: the KS system and flow over an airfoil. The datasets used for evaluation were generated using the policy obtained from the standard online SAC algorithm. For each environment, we constructed two datasets—one using a medium-performing policy and another using an expert policy.

Each KS dataset consists of 10,000 trajectories generated using its respective policy (medium or expert). For the flow over airfoil environment, each dataset comprises 2,000 trajectories, also generated using the corresponding policy. The number of trajectories in the airfoil case was intentionally limited due to the large memory footprint—approximately 1.5 TB—resulting from the high-dimensional state representations. Each trajectory in the dataset includes the full current state $\mathbf{s}_t$, the action taken $\mathbf{a}_t$, the resulting next state $\mathbf{s}_{t+1}$, and the corresponding reward $r$, providing complete transition information necessary for offline RL training.

\subsection{Extracting Policies for Various Observations}

\subsubsection*{KS environment}
In real-world applications, sensor placements and the corresponding partial observations often vary across deployments, typically requiring retraining the policy through environment interaction for each configuration. The SACN addresses this challenge by enabling the extraction of policies tailored to different sensor layouts without additional interaction. To evaluate this capability, we begin with experiments on the KS environment, where the system is partially observed using eight sensors and controlled through four actuators.

\begin{figure}[!ht]
    \centering
    \includegraphics[width=\linewidth]{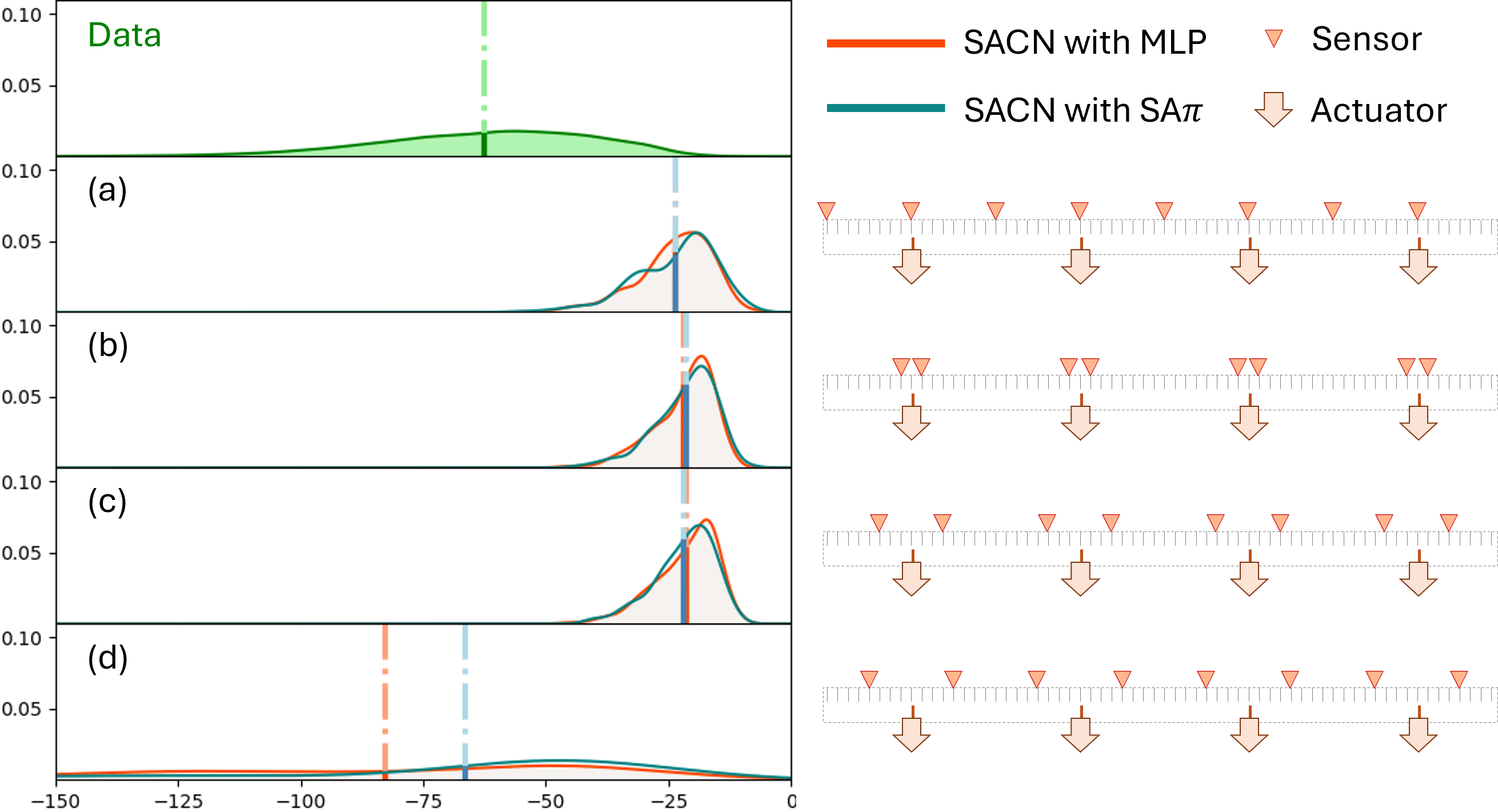}
    \caption{Comparison of return distributions between the learned policies using SACN across various sensor layouts (illustrated on the right panel) and the behavior policy used to generate the dataset in the KS environment. Policies are trained using SACN with both the standard MLP and the PC$\pi$-Net.}
    \label{fig:ks_sacn_po}
\end{figure}

In this setup, the goal is to extract four distinct policies, each corresponding to a unique partial observation configuration. These configurations—shown on the right side of Fig.~\ref{fig:ks_sacn_po}—range from sensor placements near the actuators to those farther away. Each sensor provides only a scalar measurement of the state at its location. 

We use the KS medium-policy dataset to train policies using the SACN algorithm, with sensor locations fixed during each training run. For the policy network, we consider both a standard MLP and our proposed PC$\pi$-Net, which takes partial observation as input. The Q-network is trained with the full state as input.
Four separate training sessions are conducted for each approach—one for each distinct observation configuration.
We evaluate the resulting policies on the KS environment using 200 different initial conditions. The corresponding return distributions, shown on the left side of Fig.~\ref{fig:ks_sacn_po}, compare the performance of the learned policies against that of the behavior policy used to generate the dataset. These results demonstrate that SACN can extract effective policies under varying sensor configurations in a fully offline setting using both the MLP and the PC$\pi$-Net. Both the standard MLP and the PC$\pi$-Net result in similar performance, indicating that the position-condition architecture does not degrade policy quality.
Notably, the policies corresponding to configurations (a), (b), and (c) outperform the behaviour policy used to generate the dataset, indicating successful policy improvement. In contrast, the policy learned for configuration (d) yields lower returns, which we attribute to the sensor placements being furthest from the actuators, limiting observability of the control-relevant dynamics.

\subsubsection*{Flow over airfoil environment}
\begin{figure}[!htp]
    \centering
    \includegraphics[width=0.5\linewidth]{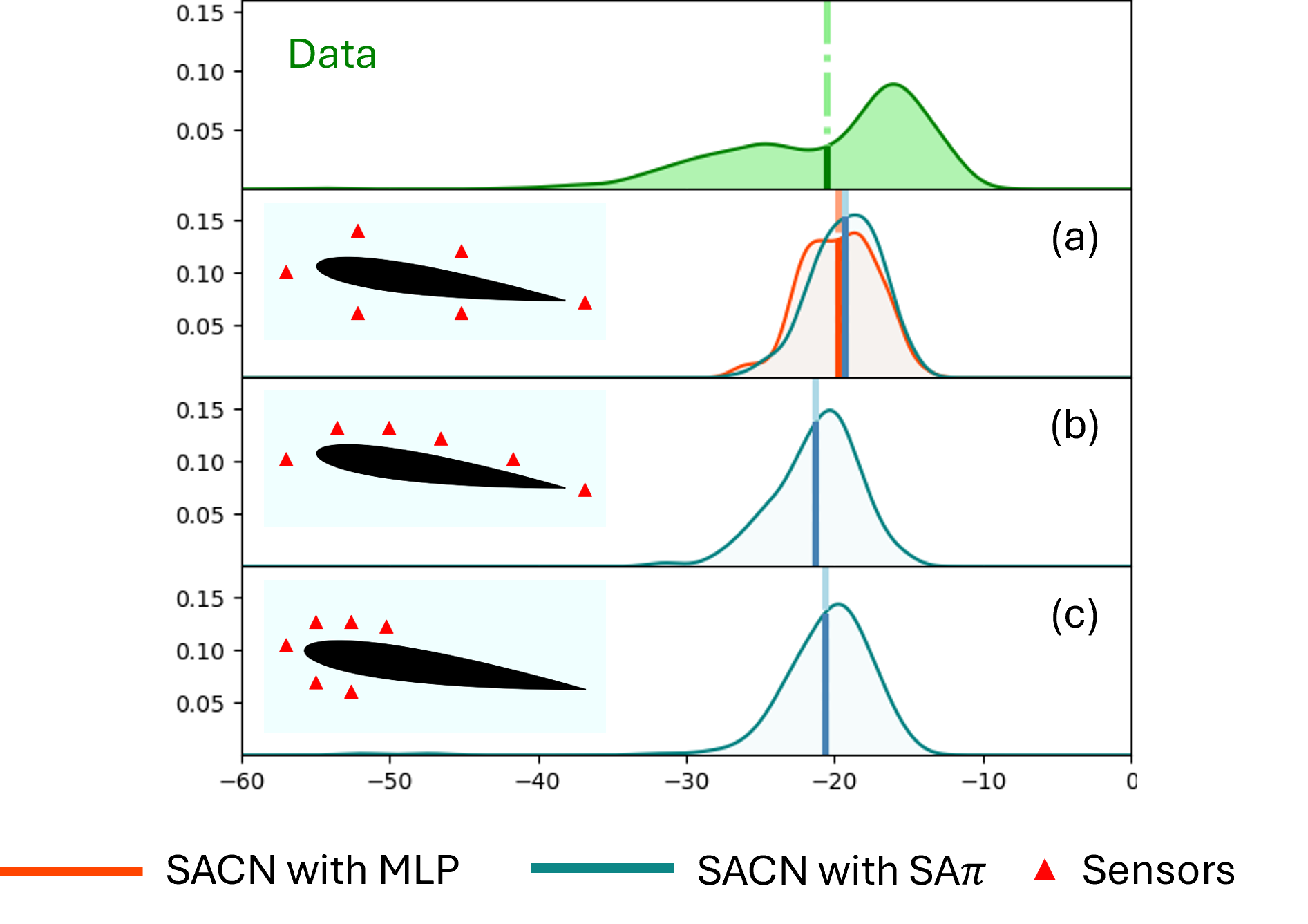}
    \caption{Comparison of return distributions between the learned policies using SACN across various sensor layouts and the behavior policy used to generate the dataset in the flow over airfoil environment.}
    \label{fig:af_sacn_po}
\end{figure}

We conduct a similar evaluation on the flow over airfoil environment to assess the generalizability of the SACN framework. In this case, the system is controlled using three actuators and observed using six sensors placed at different locations along the airfoil surface. We consider three distinct observation configurations, each defined by a different set of sensor placements, as shown on the left side of Fig.~\ref{fig:af_sacn_po} (a-c).
Similar to the KS case, we train the PC$\pi$-Net separately for each configuration using the medium-policy dataset for the airfoil environment. Each sensor provides velocity and pressure measurements at its respective location. Both the PC$\pi$-Net and the Q-network are trained using the partial observations obtained from these sensors. 
For configuration (a), we also train a standard MLP policy to confirm that the PC$\pi$-Net achieves comparable performance. This verifies that position-condition architecture does not compromise policy quality, even in more complex fluid environments. Our proposed PC$\pi$-Net is designed to tackle realistic scenario like varying sensor locations and the introduced complexity doesn't compromise the performance on simpler problem settings.  
%\todo[inline]{I didn't get the point here, what does it indicate if PC$\pi$ performs the same as MLP. What is the gain? Does it prove without sensor aware setting, the PC$\pi$-net converges to MLP?}

We deploy the learned policies on 200 test rollouts with different inlet velocity profiles and record the corresponding returns. Fig.~\ref{fig:af_sacn_po} presents the performance of the extracted policies across the three sensor configurations. The histograms show that, despite variation in sensor placement, the policies trained under different partial observations achieve performance comparable to that of the behavior policy used for data generation. This further demonstrates the flexibility of the SACN framework in handling partial observability through varied sensor layouts in complex fluid environments.

\subsection{Sensor Position-Condition Policy}
\begin{figure}[!ht]
    \centering
    \includegraphics[width=\linewidth]{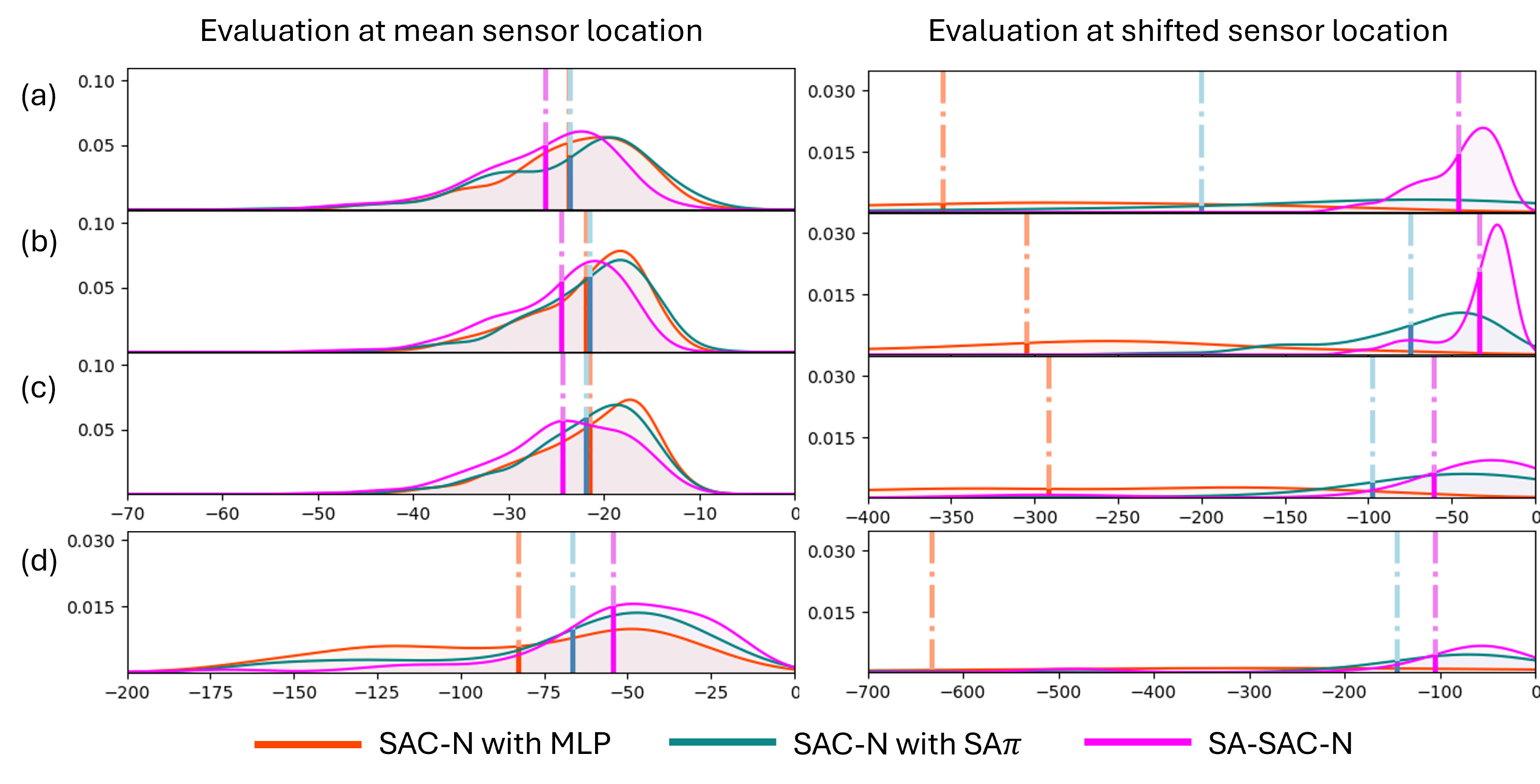}
    \caption{Comparison of return distributions for policies learned using SACN with MLP, SACN with PC$\pi$-Net, and SA-SACN across various sensor layouts in the KS environment. The left panel shows return distributions for evaluations at the respective mean sensor locations over 200 test rollouts. The right panel presents return distributions for evaluations on ten randomly perturbed sensor layouts, each tested over 20 test rollouts.}
    \label{fig:ks_multi_policy}
\end{figure}

The experiments presented thus far were conducted independently, with each policy requiring its own separately trained network. Although Offline RL eliminates the need for direct interaction with the environment during policy extraction, retraining the network whenever sensor locations shift—even slightly—can be costly or impractical. 
Moreover, applying a policy trained on one sensor configuration to a shifted configuration without adaptation typically results in poor performance.
To illustrate this, we evaluate the previously trained KS policies on sensor configurations that are slightly perturbed from their original positions. Specifically, for each configuration, ten random sensor layouts are generated by sampling from a normal distribution centred at the trained sensor positions $\{\psi_i\}_{i=1}^8$, with a standard deviation of $\sigma = 1/(3\times 16)$. Policy for each sensor layout is then evaluated on 20 different initial conditions. The resulting return distributions of PC$\pi$ trained with SACN, shown on the right side of Fig.~\ref{fig:ks_multi_policy}, demonstrate a significant drop in performance. 
We conduct a similar evaluation using the standard MLP policy. Unlike the PC$\pi$-Net, the MLP lacks a mechanism to handle continuously varying sensor locations. To approximate this behaviour, each sampled sensor position is rounded to the nearest value in the discrete set $\{ i/64 \}_{i=0}^{63}$, and the corresponding observation is used for inference. As shown in Fig.~\ref{fig:ks_multi_policy}, the MLP-based policy performs even worse under sensor perturbation, highlighting its inability to generalize across sensor configurations.

To address this limitation, we trained the PC$\pi$-Net with an SA-SACN algorithm to learn policies corresponding to a range of sensor locations centred around a mean placement $\{\psi_i\}_{i=1}^8$. Specifically, for every epoch during training, sensor locations are sampled as $ \textbf{x}_i \sim \psi_i + (2\mathcal{U}(\cdot)-1)\Delta$. This multi-observational training strategy enables the policy to generalize across sensor variations, allowing sensor positions to be adjusted at deployment without requiring retraining.
We conduct four training sessions using different mean sensor configurations (see right side of Fig.~\ref{fig:ks_sacn_po}) with $\Delta = 1/(3\times 8)$, all based on the KS medium-policy dataset. Figure~\ref{fig:ks_multi_policy} compares the performance of policies trained using single fixed configurations with SACN versus those trained using the SA-SACN. The left panel shows the return distributions for policies evaluated at their respective mean sensor locations over 200 test rollouts with varying initial conditions. The right panel presents the return distributions for policies evaluated on ten randomly perturbed sensor layouts, each tested over 20 different initial conditions. These sensor layouts are generated by sampling from a normal distribution centered at the trained sensor positions $\{\psi_i\}_{i=1}^8$, with a standard deviation of $\sigma = 1/(3\times 16)$. The results demonstrate that SA-SACN significantly improves the robustness to sensor variability.

\begin{figure}[!htp]
    \centering
    \includegraphics[width=0.95\linewidth]{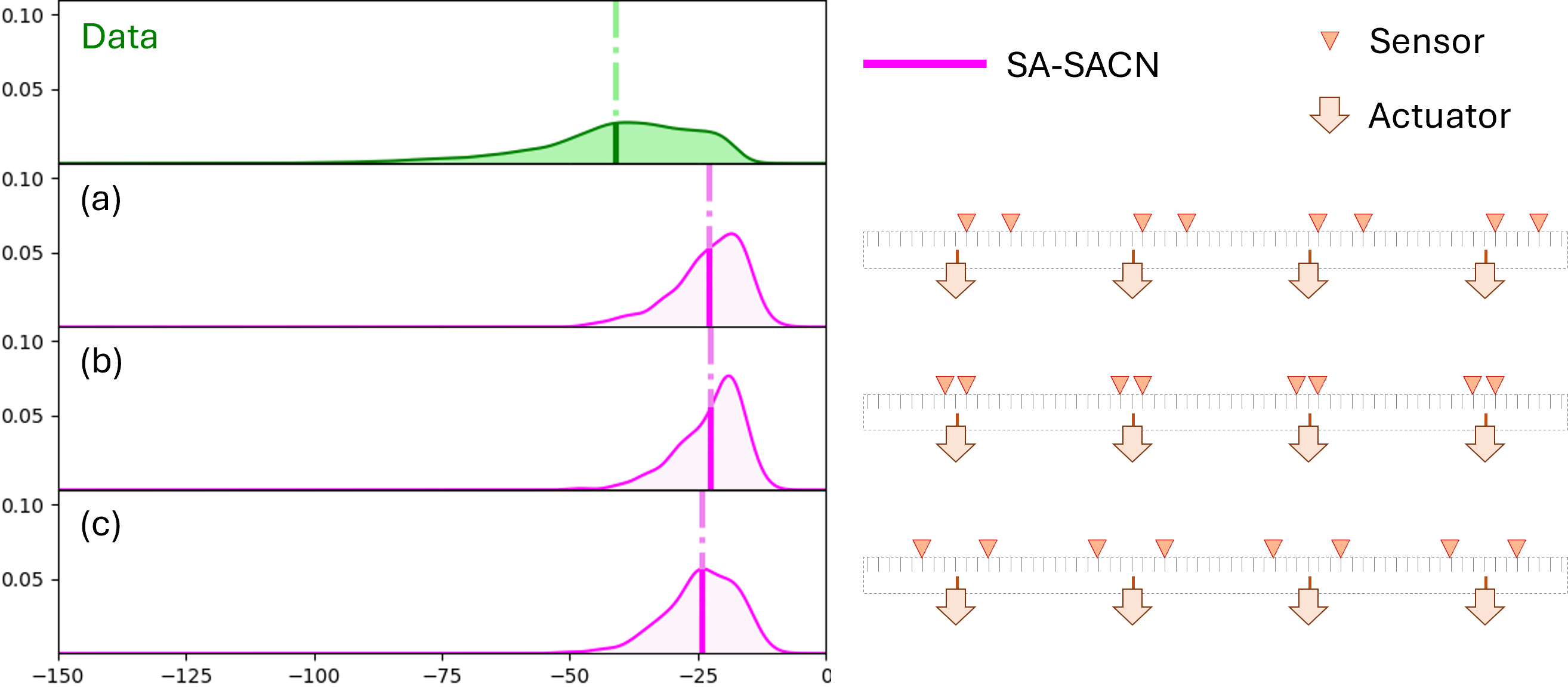}
    \caption{Comparison of return distributions for policies learned using SA-SACN with configuration (c) from Fig.~\ref{fig:ks_sacn_po} as the mean sensor layout, evaluated across various sensor configurations (illustrated on the right panel) in the KS environment.}
    \label{fig:ks_multi_policy1}
\end{figure}
Additionally, the PC$\pi$-Net trained for configuration (c) (see Fig.~\ref{fig:ks_sacn_po}) is tested on other configurations, and the return distributions are shown in Fig.~\ref{fig:ks_multi_policy1}. This clearly shows performance of the PC$\pi$-Net with the SA-SACN algorithm captures various policies corresponding to various sensor locations.

\begin{figure}[!htp]
    \centering
    \includegraphics[width=\linewidth]{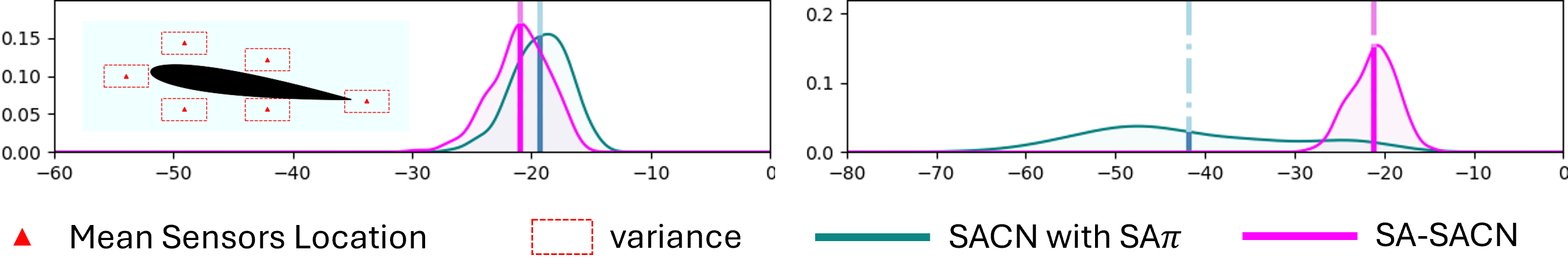}
    \caption{Comparison of return distributions for policies learned using SACN with MLP, SACN with PC$\pi$-Net, and SA-SACN in the flow over airfoil environment. The left panel shows return distributions for evaluations at the mean sensor locations over 200 test rollouts. The right panel presents return distributions for evaluations on ten randomly perturbed sensor layouts, each tested over 20 test rollouts.}
    \label{fig:af_multi_policy}
\end{figure}

A similar evaluation on the flow over an airfoil environment to test the generalization capability of the PC$\pi$-Net with SA-SACN under sensor location shifts was conducted. The PC$\pi$-Net is trained with a mean placement $\{\psi_i\}_{i=1}^6$ with $\Delta=1/(3\times 10)$ along the $x$-axis and $\Delta=1/(3\times 20)$ along the $y$-axis.
Fig.~\ref{fig:af_multi_policy} compares the performance of policies trained using single fixed configurations using SACN versus those trained using the SA-SACN. The left side of the figure shows the return distributions for the policies evaluated at mean sensor locations, while the right side shows the distributions for ten random sensor layouts that are generated by sampling from a normal distribution centred at the trained sensor positions $\{\psi_i\}_{i=1}^6$, with a standard deviation of $1/(3\times 20)$ and $1/(3\times 40)$ along $x$ and $y$ axis, respectively. The return distributions, shown in Fig.~\ref{fig:af_multi_policy}, confirm consistent findings: the SA-SACN maintains strong performance across varying sensor layouts without requiring retraining, demonstrating its robustness and adaptability in more complex fluid environments.

%% file: discussion.tex
\section{Discussion}
\label{sec:Discussion}
\subsection{Sensor Optimization}
\begin{figure}[!htp]
    \centering
    \includegraphics[width=\linewidth]{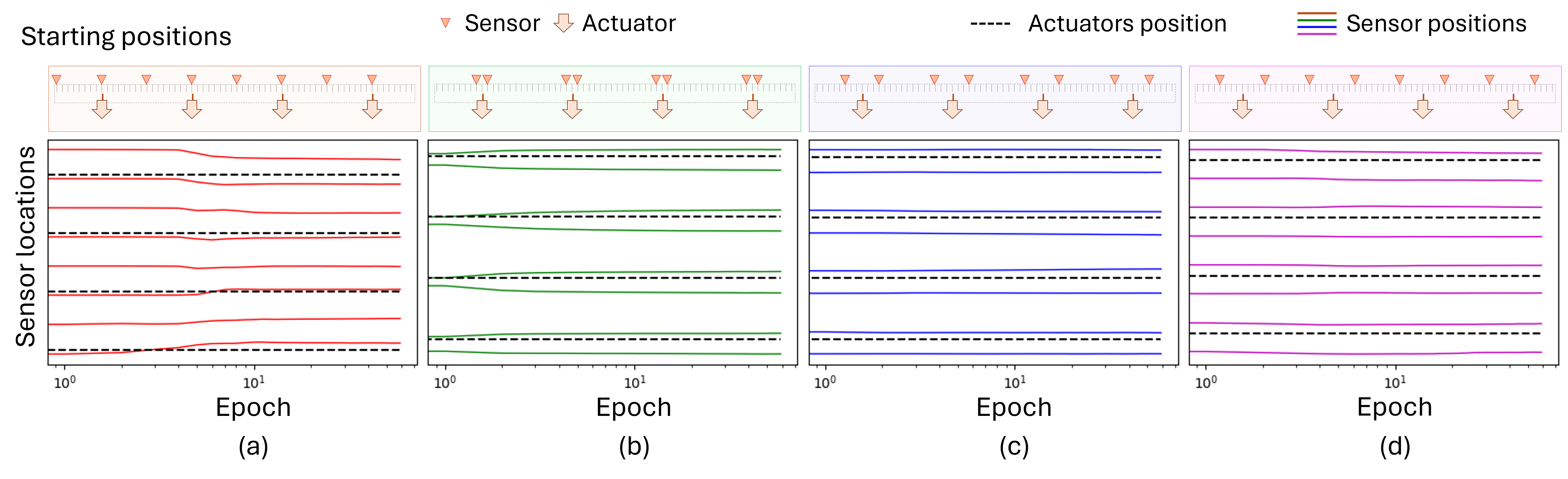}
    \caption{Sensor location optimization over training epochs across various initial sensor layouts in the KS environment. The top panel illustrates the initial sensor layouts.}
    \label{fig:sen_opt}
\end{figure}

Given that the PC$\pi$-Net, combined with an PC-SACN algorithm, has demonstrated the ability to learn multiple policies corresponding to different sensor configurations, it can also be leveraged to identify optimal sensor placements that yield the best policy performance. As illustrated in Fig.~\ref{fig:ks_sacn_po}, in the KS environment, certain configurations such as (a), (b), and (c) result in higher-performing policies, whereas others like (d) lead to significantly worse outcomes.
To explore this further, the PC$\pi$-Net was trained across all four configurations shown in Fig.~\ref{fig:ks_sacn_po} using the KS medium-policy dataset. In addition to the PC-SACN training objective, a sensor optimization objective (defined in Eq.~\ref{eq:sen_opt_obj} was incorporated to adjust sensor placements $\{\psi_i\}_{i=1}^8$ during training. The evolution of sensor positions over training epochs is shown in Fig.~\ref{fig:sen_opt}. While the optimized sensor configurations do not converge to a unique layout, all configurations tend to shift toward a pattern similar to that of configuration (c).
% \textcolor{red}{in Fig~\ref{fig:ks_sacn_po}}, 
% \todo[inline]{It is more clear if we could put the configureation (c) in Fig 5 in Fig 10. It is ambiguous if you mean config c in Fig 10 or Fig 5. Then if Fig 10 looks too narrow, maybe we could remove one of the existing configs, so we will have 3 different initial positions and one coverged config positions pasted from previous result?}
% which previously yielded strong performance.
%\todo[inline]{I didn't see the grey "mean line" in the Figure 11.}
\begin{figure}[!ht]
    \centering
    \includegraphics[width=0.95\linewidth]{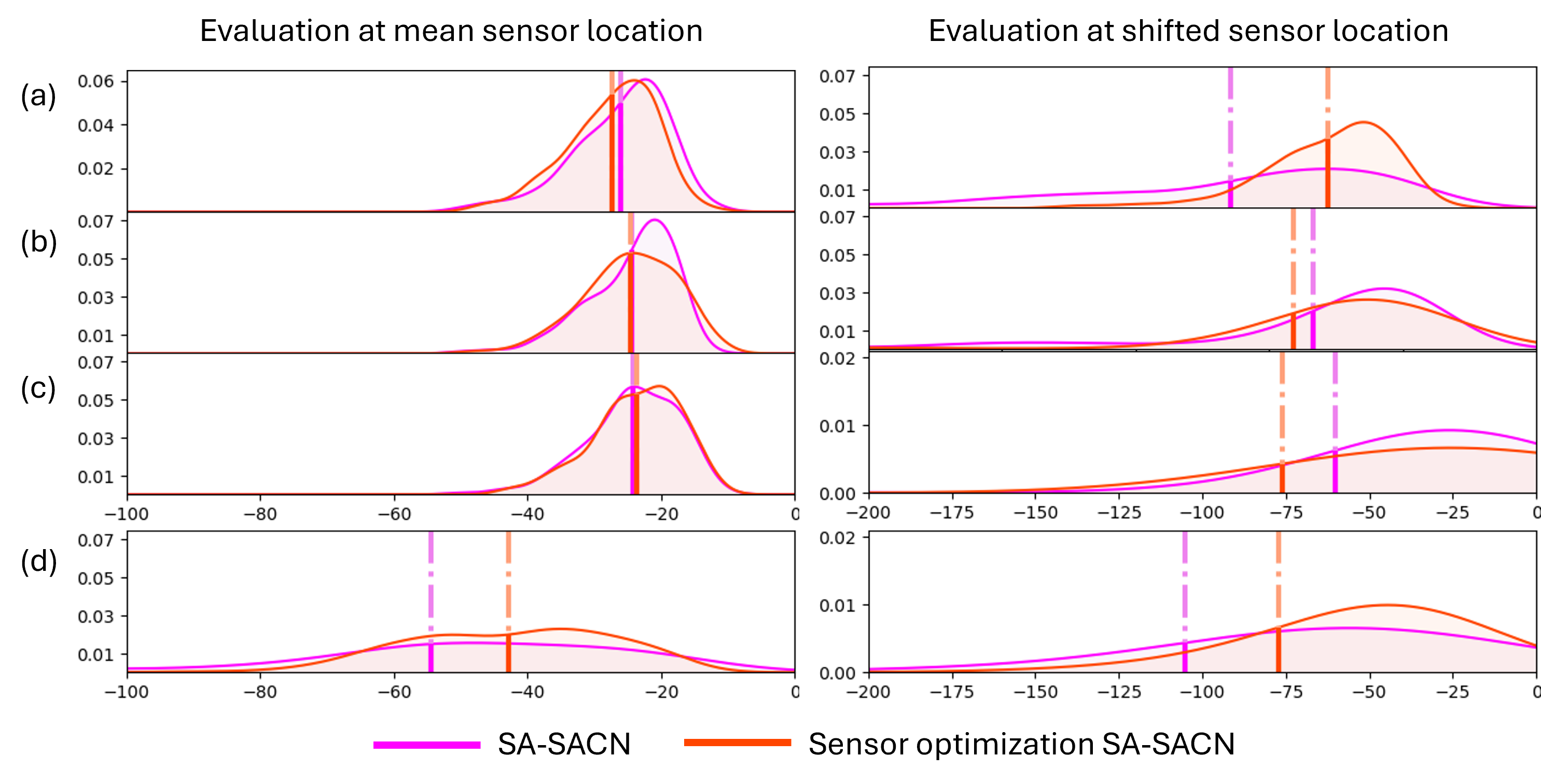}
    \caption{Comparison of return distributions for policies learned using PC-SACN and PC-SACN with sensor optimisation objective across various initial sensor layouts in the KS environment. The left panel shows return distributions for evaluations at the respective mean sensor locations over 200 test rollouts. The right panel presents return distributions for evaluations on ten randomly perturbed sensor layouts, each tested over 20 test rollouts.
    KS offline PC-SACN robust performance for various sensor locations}
    \label{fig:ks_sen_opt_returns}
\end{figure}

The return distributions for the optimized sensor layouts are presented in Fig.~\ref{fig:ks_sen_opt_returns}. 
% \textcolor{red}{LS:The left panel shows the distribution at the mean sensor location, as marked by "Sensors" in Fig.~\ref{fig:af_multi_policy}. And the right panel shows the result for the shifted sensor locations, where the perturbation is generated based on the mean and variance as sketched in Fig.~\ref{fig:af_multi_policy}. We compared the performance with and without sensor optimization using PC-SACN with PC$\pi$ block.
% } \textcolor{red}{In the evaluation at mean sensor scenario} 
for cases (b) and (c) , the performance doesn't improve as the sensor positions are already the best; however for cases (a) and (d), the optimised sensor position results in better performance. Furthermore, in evaluation at shifted sensor location, the sensor optimization results are better for all four initial layouts, which shows that the sensor optimization objective function learnt more robust policies with respect to various sensor locations.
This demonstrates the potential of the proposed framework to identify sensor placements that enable optimal policy performance, solely based on the available offline dataset.
% \todo[inline]{could you remind me why no = optimization for airfoil? Is it too expensive?}

%% file: conclusion.tex
\section{Conclusion}
\label{sec:Conclusion}

This work presents a novel offline Reinforcement Learning framework for fluid control tasks, addressing the challenges associated with varying sensor configurations and the high cost of environment interactions. By leveraging pre-collected datasets and the SACN algorithm, we demonstrate the ability to extract effective control policies for complex fluid environments, including the Kuramoto–Sivashinsky system and flow over an airfoil.

A key contribution of this work is the SA$\pi$-Net, which enables a single policy model to generalize across multiple observation configurations. Coupled with an PC-SACN training algorithm, this network eliminates the need for retraining when sensor placements change, significantly reducing deployment overhead. Furthermore, we introduce a sensor optimisation objective that allows the model to learn not only the policy but also the most informative sensor placements, leading to improved control performance. 
% \todo[inline]{Some reviewer may ask about the speed up ratio. If you have result ,add it. If not, be prepared to do it during rebutaal.}

Experimental results validate the proposed approach across both benchmark environments, showing robustness to sensor variability and competitive performance compared to policies trained on fixed configurations. The proposed framework offers a promising direction for scalable, data-driven control in physical systems where sensor flexibility and data efficiency are critical.

Future work will focus on improving training efficiency, as learning from high-dimensional fluid datasets remains computationally demanding. Additionally, we plan to extend the framework to more complex 3D flow environments and explore real-time deployment in experimental setups.